\documentclass[pdflatex,sn-mathphys-num]{sn-jnl}
\usepackage{graphicx}%
\usepackage{multirow}%
\usepackage{amsmath,amssymb,amsfonts}%
\usepackage{amsthm}%
\usepackage{mathrsfs}%
\usepackage[title]{appendix}%
\usepackage{xcolor}%
\usepackage{textcomp}%
\usepackage{manyfoot}%
\usepackage{booktabs}%
\usepackage{algorithm}%
\usepackage{algorithmicx}%
\usepackage{algpseudocode}%
\usepackage{listings}%
\usepackage{url}
\usepackage{tabularx}
\usepackage{subcaption}
\usepackage{booktabs}
\usepackage{makecell}
\usepackage[normalem]{ulem}
\usepackage[section]{placeins}

\def\BibTeX{{\rm B\kern-.05em{\sc i\kern-.025em b}\kern-.08em
    T\kern-.1667em\lower.7ex\hbox{E}\kern-.125emX}}
\useunder{\uline}{\ul}{}

\begin{document}

\title[Spiking Neural Networks for Radio Frequency Interference Detection in Radio Astronomy]{Spiking Neural Networks for Radio Frequency Interference Detection in Radio Astronomy}

\author*[1, 2]{\fnm{Nicholas J.} \sur{Pritchard}}\email{nicholas.pritchard@research.uwa.edu.au}
\author[1]{\fnm{Andreas} \sur{Wicenec}}\email{andreas.wicenec@icrar.org}
\equalcont{These authors contributed equally to this work.}
\author[2]{\fnm{Mohammed} \sur{Bennamoun}}\email{mohammed.bennamoun@uwa.edu.au}
\equalcont{These authors contributed equally to this work.}
\author[1]{\fnm{Richard} \sur{Dodson}}\email{richard.dodson@icrar.org}
\equalcont{These authors contributed equally to this work.}
\affil*[1]{\orgdiv{International Centre for Radio Astronomy Research}, \orgname{University of Western Australia}, \orgaddress{\street{7 Fairway}, \city{Perth}, \postcode{6009}, \state{WA}, \country{Australia}}}
\affil[2]{\orgdiv{School of Physics, Mathematics and Computing}, \orgname{University of Western Australia}, \orgaddress{\street{35 Stirling Highway}, \city{Perth}, \postcode{6009}, \state{WA}, \country{Australia}}}

\abstract{
Spiking Neural Networks (SNNs) promise efficient and dynamic spatio-temporal data processing.
This paper reformulates a significant challenge in radio astronomy, Radio Frequency Interference (RFI) detection, as a time-series segmentation task suited for SNN execution. 
Automated systems capable of real-time operation with minimal energy consumption are increasingly important in modern radio telescopes.
We explore several spectrogram encoding methods and network parameters, applying first and second-order leaky integrate and fire SNNs to tackle RFI detection.
We introduce a divisive normalisation-inspired pre-processing step, improving detection performance across multiple encodings strategies.
Our approach achieves competitive performance on a synthetic dataset and compelling initial results on real data from the Low-Frequency Array (LOFAR).
We position SNNs as a viable path towards real-time RFI detection, with many possibilities for follow-up studies.
These findings highlight the potential for SNNs performing complex time-series tasks, paving the way towards efficient, real-time processing in radio astronomy and other data-intensive fields.
}
\maketitle
\section*{Introduction}
Spiking Neural Networks (SNNs) are inspired by how biological neurons communicate.
Instead of using continuous signals, as in Artificial Neural Networks (ANNs), they transmit information in discrete spikes or pulses, closer to how natural neurons fire \cite{trappenberg_fundamentals_2010}.
ANNs have proven to be highly effective, especially when deployed on conventional computing hardware platforms, and they excel in numerous tasks when combined with large datasets.
The central question remains whether SNNs can reliably achieve comparable or superior results by mimicking biological neurons more closely while using less energy and computational power, particularly when running on neuromorphic computing systems \cite{ottati_spike_2023}.
The potential for SNNs to outperform traditional models hinges on developing efficient neuromorphic hardware and identifying data-intensive applications that can exploit the temporal dynamics unique to SNNs \cite{schuman_opportunities_2022, muir_road_2025}.
While SNNs have shown promise in several domains, particularly those leveraging neuromorphic sensing hardware, tackling increasingly demanding tasks is vital to moving the field forward.

Radio astronomy involves coordinating increasingly complex observatories and massive computing facilities to process the increasingly vast amount of data generated by observing the cosmos \cite{vermij_challenges_2015}.
One of the primary challenges in this field is dealing with Radio Frequency Interference (RFI), which refers to unwanted radio signals originating from human-made or terrestrial sources.
RFI can significantly contaminate radio astronomy data, and detecting these polluted regions in observed spectrograms and subsequently removing them from further analysis is crucial for preserving the scientific integrity of the observations.
The need for adaptive, data-driven RFI detection methods grows as radio telescopes become more sensitive and RFI becomes more pervasive.
While many factors contribute to increased RFI prevalence, the sudden increase in low-Earth orbit satellites operating over more of the radio spectrum is chief among them, even in sparsely populated parts of the Earth where terrestrial sources are less prevalent \cite{noauthor_iau_2019, noauthor_statement_2019, noauthor_report_2022}.
Traditional approaches to RFI detection often rely on cumulative-sum algorithms fine-tuned to specific instruments and environments \cite{offringa_aoflagger_2010}, in combination with solutions built into the interferometry hardware \cite{perez-portero_rfi_2022, mohamed-fakier_transient_2024}.
Contemporary experimental machine learning techniques based on UNet-like CNNs \cite{akeret_radio_2017, vafaeisadr_deep_2020, yang_deep_2020}, anomaly detection schemes \cite{mesarcik_learning_2022, vanzyl_remove_2024} or vision-transformer models \cite{ouyang_hierarchical_2024} offer competitive, and sometimes superior, results, albeit with high operational costs and a need for extensive training data \cite{dutoit_comparison_2024}.

Both conventional and machine-learned approaches treat RFI detection as a two-dimensional semantic segmentation or anomaly detection problem, where spectrograms, or `visibilities,' are treated like images.
Although effective, existing two-dimensional approaches to RFI detection necessitate completing an entire observation before processing can commence.
This is a significant limitation as data destined to be removed is stored and processed in costly high-performance computing facilities \cite{vermij_challenges_2015}.
As such, there is a strong imperative in the field to develop RFI detection techniques that are accurate and capable of operating in near real-time, with minimal energy consumption, and where applicable, minimal training data \cite{mesarcik_learning_2022, dutoit_comparison_2024}.
Previous works have investigated real-time RFI detection surveying common mitigation strategies at various observatories \cite{buch_real-time_2025}, developing scalable real-time flaggers for large-scale arrays such as LOFAR \cite{van_nieuwpoort_towards_2016} and the Apertif Radio Transient System (ARTS) at the Westerbork Synthesis Radio Telescope (WSRT) \cite{sclocco_real-time_2019}, and investigating the precise effects of persistent but still dynamic satellite sources and their effects on precision science experiments \cite{engelbrecht_radio_2025}.

The time-varying nature of radio astronomy data and its spectrographic characteristics suggest that this domain could benefit from SNNs and neuromorphic computing \cite{kasabov_evolving_2016}.
Despite this potential, only a few attempts have been made to apply SNNs within radio astronomy \cite{scott_evolving_2015}.
More specifically, regarding RFI detection, only one existing study employs SNNs \cite{pritchard_rfi_2024}.
This work utilised ANN-to-SNN conversion in an auto-encoder-based anomaly detection framework.
While it demonstrated operational benefits, the approach did not fully take advantage of the inherent time-varying aspects of radio astronomy observations, leaving much of the potential for SNNs untapped in this area.

Previous applications of SNNs to spectrographic data, such as audio processing, have primarily focused on classification tasks \cite{stewart_speech2spikes_2023, bos_sub-mw_2022}.
In such tasks, a time-varying input is processed, but the final output is reduced to a single value or decision derived from the final layer of the network.
However, the problem of RFI detection requires the output to retain the same temporal and spatial resolution as the input data.
In this work, we reformulate RFI detection as a time-series segmentation problem, where the network must output a prediction that spans the same time and frequency dimensions as the input spectrogram to explore how well first and second-order leaky-integrate and fire (LiF) spiking neurons can capture temporal information in this task.
The comparison between neuron types is also significant, as second order neurons are often more difficult to train. We report on the best-performing neuron type for each dataset in the main-text, and include the remaining results in Supplementary Notes 1 and 2.
To our knowledge, this is the first time SNNs have been applied to this kind of task, and until now, it has been unknown whether backpropagation through time (BPTT) can handle the complexity of such tasks.

Transitioning from synthetic datasets to real radio astronomy data introduces a significant increase in noise and variability.
To assist in handling this challenge, we introduce a pre-processing technique inspired by divisive normalisation, a sensory adaptation mechanism rooted in neuroscience \cite{carandini_linearity_1997}.
Divisive normalisation has been studied extensively and incorporated into image segmentation networks \cite{hernandez-camara_neural_2023}, and we have adapted the basic concepts into a spectrogram pre-processing step.
This pre-processing technique relies only on values from a single preceding time step, effectively increases the contrast between RFI and background signals and, critically, applies universally to all downstream spike-encodings.
This technique increases spike sparsity in all encodings and significantly improves performance on the synthetic and real radio astronomy datasets we investigate in this work.

This paper provides a robust formulation of RFI as a time-series segmentation task appropriate for direct SNN execution, investigates encoding visibility data with latency, rate, delta-modulation, delta-exposure and three distinct variations on step-forward encoding, extensive hyper-parameter optimisation, a divisive normalisation inspired pre-processing step which greatly improves detection performance across several encoding methods, and results for SNNs trained and optimised with each encoding for a synthetic dataset simulating the Hydrogen Epoch of Reionisation Array (HERA) in South Africa, and a real RFI detection dataset derived from the Low-Frequency Array (LOFAR) instrument in the Netherlands with comparison to identically sized ANN feed-forward networks.
We opt to compare results to directly sized, very simple ANNs to highlight the specific differences introduced by using spiking neurons specifically.
This paper builds upon our previously presented work \cite{pritchard_supervised_2024}, extending it in several important ways.
The delta-exposure encoding method and divisive-normalisation-inspired pre-processing step are previously unseen techniques.
We provide more comprehensive hyper-parameter optimisation extending into network depth and width in addition to a more extensive analysis of the resulting parameter space.
Lastly, by including results for the real LOFAR dataset, we present the first study, to the best of our knowledge, demonstrating the application of from-scratch trained SNNs on real radio astronomy data.
In doing so, we achieve state-of-the-art RFI detection performance on the synthetic HERA dataset and initial results on the real LOFAR dataset, demonstrating the efficacy of our approach while employing small multi-layer feed-forward SNNs.
This investigation demonstrates the capability of SNNs to perform a data-intensive complex time-series segmentation task in the form of RFI detection with relatively simple network architectures, paving the way for more sophisticated SNN-based RFI detection schemes and critically finding a scientifically impactful problem in which SNNs and neuromorphic computing may hold unique advantages over contemporary methods.
\section*{Results}
\subsection*{RFI Detection as Time-Series Segmentation}
The signal chain in a radio interferometer array observatory, generally speaking, first involves correlating raw voltage signals from antennae, transforming the data into complex-valued `visibilities' $V(\upsilon, T, b)$, which vary in frequency, time and baseline (pair of antennae) \cite{thompson_interferometry_2017}. 
At this stage in the processing chain, RFI detection involves producing a boolean mask of `flags' $G(\upsilon, T, b)$ that vary in the same parameters but are binary-valued.
These binary masks are used in subsequent signal processing steps to speed up the collation of visibilities into image cubes which are then eventually handed off to astronomers to begin scientific processing. Increasing the throughput of RFI detection directly improves the throughput of data to end-user scientists.

Previous supervised approaches to RFI detection via machine learning formulate the problem as
\begin{equation}\label{eq:originalformulation}
    \mathcal{L}_{sup} = 
    min_{\theta_n}\mathcal{H}(
        m_{\theta_n}(
            V(\upsilon, T, b)
        ), 
        G(\upsilon, T, b)
    ),
\end{equation}
where $\theta_n$ are the parameters of some classifier $m$ and $\mathcal{H}$ is an entropy-based similarity measure \cite{mesarcik_learning_2022}.
We exploit the time-varying nature of this information and, therefore, include additional spike encoding and decoding steps and push the similarity measure inwards, operating on each element of the resulting time-series.
\begin{equation}\label{eq:newformulation}
    \mathcal{L}_{sup} = 
    min_{\theta_n}(
        \Sigma_t^T
        \mathcal{H}(
            m_{\theta_n}(
                E(V(\upsilon, t, b))
            ), 
            F(G(\upsilon, t, b))
        )
    )
\end{equation}
where $E$ is an input encoding function and $F$ is an output encoding function. Both functions introduce a new integer exposure parameter, controlling the spike train length for each time step in the original spectrogram.

This formulation tasks a classifier to segment the time series, in this case, as boolean values rather than a two-dimensional image.
In the case of an ANN, the encoding and decoding functions are the identity.
This formulation leaves encoding and decoding visibility data as spikes for this specific task open for experimentation.
The spike encoding process involves stretching the original time steps in the spectrograms into longer exposures, allowing the SNN to react to the current input. 
We experimented with seven encoding methods.
The methods section presents detailed formulations and examples of each encoding approach. 

\begin{figure*}[!htbp]
    \centering
    \includegraphics[width=\textwidth, keepaspectratio]{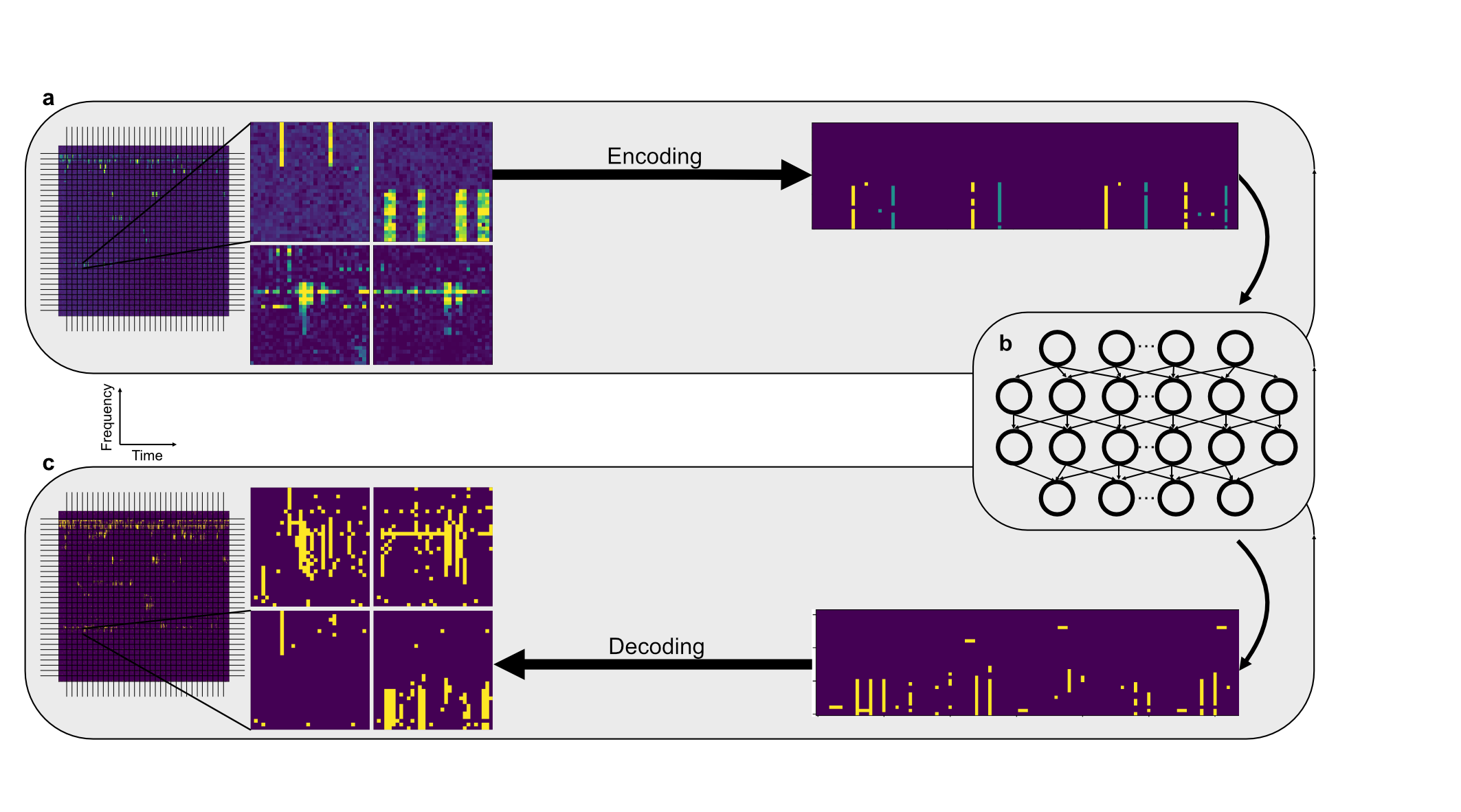}
    \caption{Spiking Neural Network (SNN) Based Radio Frequency Interference (RFI) Detection Workflow. (a) Encoding: A spectrogram is divided into smaller patches and encoded into spikes using methods like delta-exposure encoding (shown with an exposure time of four).
    (b) SNN Processing: The encoded spikes are input into an SNN with a structure matching the spectrogram's frequency channels. The SNN identifies RFI patterns from the time-varying data.
    (c) Decoding: Output spikes are decoded to produce an RFI mask for each patch, which is reassembled into a complete spectrogram RFI mask.}
    \label{fig:main}
\end{figure*}

Figure \ref{fig:main} outlines our complete approach to supervised RFI detection with SNNs using delta-exposure encoding as an example.
Notably, the input width of the SNN either exactly matches or doubles (in the case of step-forward encodings) the number of input frequency channels ($32$ in this case).
For training purposes, we slice each spectrogram into $32 \times 32$ pixel patches and encode each into a spike train using the swappable encoding method.
Without patching, training an SNN with BPTT becomes very challenging as the memory requirements grows proportional to the length of the input spike-train.
We discuss difficulties introduced by patching the inputs in the Methods section.
Bounding the number of time-steps allows for tractable learning as the network only needs to learn coherent behaviour for a limited amount of time.
The entire spike train is fed into the SNN along the time axis, and supervision is performed on the SNN's output spike train.
At inference and final testing, the SNN's output spike train is then decoded into another patch that is then re-stitched into a complete spectrogram for accuracy testing.
This task is challenging for an SNN as we present each original time step in the spectrogram precisely once; identifying features that span the entire spike train requires the network to retain some of this information in the leaky dynamics of the neurons themselves; these results represent a baseline level of performance given what is as lightweight a network as possible for this task.
\FloatBarrier
\subsection*{Encoding independent divisive normalisation}
\begin{figure*}[!htbp]
        \centering
    \begin{subfigure}{0.45\columnwidth}
        \centering
        \includegraphics[width=\textwidth, keepaspectratio]{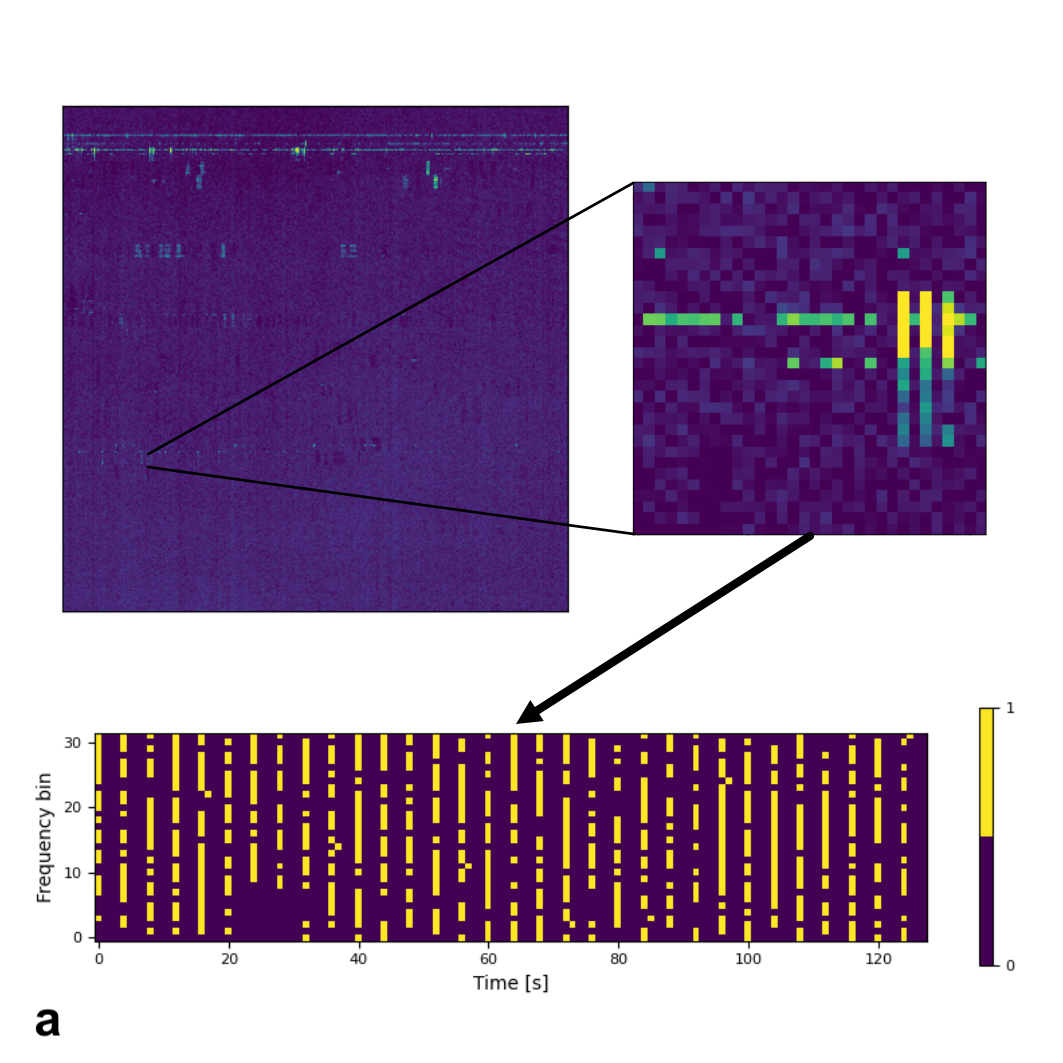}
        \label{fig:divnorm:enc:latency}
    \end{subfigure}
    \begin{subfigure}{0.45\columnwidth}
        \centering
        \includegraphics[width=\textwidth, keepaspectratio]{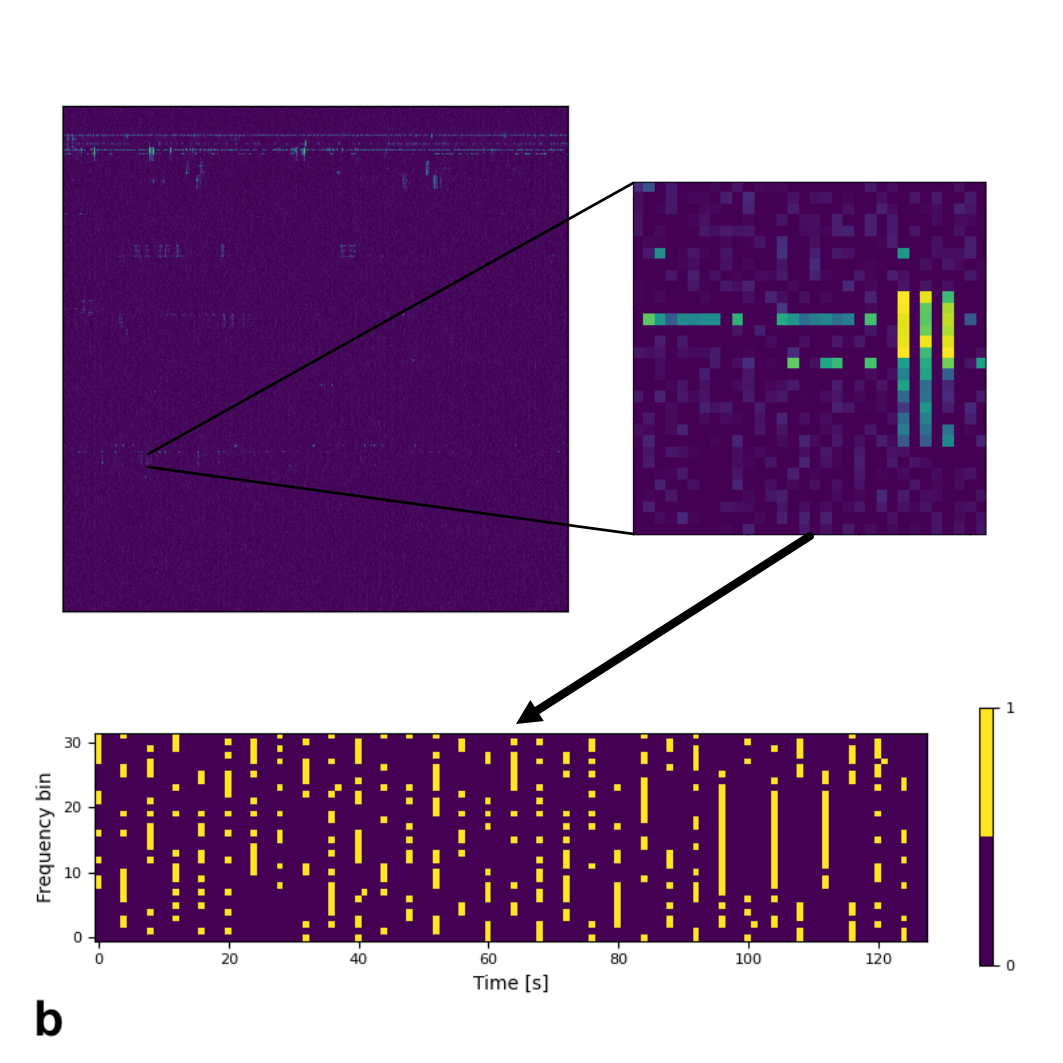}
        \label{fig:divnorm:enc:rate}
    \end{subfigure}
    \caption{Effects of divisive normalisation on spectrogram pre-processing for Radio Frequency Interference (RFI) detection.
    Panel (a) depicts an original full LOFAR spectrogram (left) and sub-section patch (right) encoded with latency encoding spike raster (bottom, four time step exposure). RFI features are difficult to distinguish from background information.
    Panel (b) depicts the same full spectrogram (left), sub-section patch (right) after divisive normalisation with a latency encoding spike raster (bottom, four time step exposure). The background gradient is reduced, increasing spike sparsity and making RFI features more prominent, like the three bars on the right.
    }
    \label{fig:divnormdemo}
\end{figure*}

Divisive normalisation is an investigated and potentially canonical neural mechanism for filtering noisy signals in brains where the activity of a neuron is modulated by the summed activity of a pool of neighbouring neurons \cite{carandini_linearity_1997, carandini_normalization_2012}.
Divisive normalisation has also been shown to play a role in choice behaviour \cite{webb_divisive_2020}.
Divisive normalisation has recently improved image segmentation performance in U-Net ANNs \cite{hernandez-camara_neural_2023}.
Knowing that most astronomical signals present as unstructured noise in a single spectrogram, we investigated applying some form of divisive normalisation to both datasets investigated.
We additionally required a technique applied to any spike encoding and, as such, formulate a divisive normalisation-like preprocessing step.
For visibility data $V(\upsilon, T, b)$, we define divisible normalised visibilities as:
\begin{equation}
    V_{dn}(\upsilon, T, b) = V(\upsilon,T,b) - \sum_{i=-\frac{k}{2}}^{i=\frac{k}{2}}V(\upsilon+i, t-1,b)
\end{equation}
where $k$ is a configurable kernel size, in our case, three, $0 \leq \upsilon+i \leq 512$ and $T-1 \geq 0$; the bounds of the frequency channels and time in the original spectrogram.
This method effectively acts as a kernel over the localised frequency channels in the immediately preceding time step.
We chose to include information from only a single previous time step to keep this method lightweight and plausible in a real-time collection setting.
Figure \ref{fig:divnormdemo} shows an example of this process acting on a LOFAR spectrogram with latency encoding.
The visibly less noisy background retains most RFI features and results in significantly fewer spikes in the encoded spike train input, effectively increasing the information carried by each spike toward the goal of RFI detection.
\subsection*{Hyper-parameter tuning on synthetic HERA dataset}
In addition to experimenting with different encoding methods, we wanted to experiment with network sizes and neuron parameters through hyper-parameter optimisation.
The HERA dataset, while artificial, contains simulated RFI sources that reflect real-world sources.
Full details of this dataset are available in the methods section.
For all trials, we measure the Area Under the Receiver Operating Characteristic curve (AUROC), the Area Under the Precision-Recall Curve (AUPRC), per-pixel Accuracy, and the F1-Score.
The definitions and significance of these values are found in the methods section.
Owing to the relatively low per-pixel prevalence of RFI in both datasets, we expect high accuracy and AUROC values but pay greater attention to AUPRC and F1-Score, class-balanced performance measures.
Our previous investigation into encoding methods tuned a restricted set of parameters, fixing the network architecture but varying batch size and training epochs. We subsequently validated the parameters discovered with repeat trials and presented the results at an earlier conference venue \cite{pritchard_supervised_2024}. 
We summarise our findings of this previous work below.
We found our SNN approach to outperform equivalent ANN networks, and informed our decision to focus on delta-exposure, latency and step-forward direct encodings in the new work presented in this article.

Here, we present results from the search through additional network shaping parameters like layer depth and width and spiking neuron parameters.
The methods section contains full details of the parameter space searched.
Table \ref{tab:hera:optuna} contains results from our hyper-parameter search exploring network parameters for delta-exposure, latency and step-forward-direct encoding method with and without divisive normalisation on the synthetic HERA dataset.
We find that latency encoding with divisive normalisation performs best in all metrics except AUROC, which is narrowly beaten out by a raw ANN with divisive normalisation.
The optimiser generally favoured larger than 128-sized hidden layers and increased network depth.
Delta exposure encoding favoured higher beta values and larger exposure times, indicating a preference for fast decaying neurons, with latency and step-forward encodings preferring shorter exposure times and slower decaying neurons.
\begin{sidewaystable}[!htbp]
\caption{Hyper-parameter search using Optuna multi-variate optimisation for the synthetic HERA dataset.}
\label{tab:hera:optuna}
\begin{tabular}{ccccccccc}
\hline
Encoding Method               & Accuracy       & AUROC          & AUPRC          & F1-Score       & \makecell{Num\\Hidden} & \makecell{Num\\Layers} & Beta       & Exposure   \\ \hline
Delta-Exposure                &                &                &                &                &            &        &            &            \\
                              & 0.990          & 0.864          & 0.791          & 0.764          & 512        & 3      & 0.435      & 53         \\
                              & 0.990          & 0.885          & 0.787          & 0.767          & 512        & 4      & 0.621      & 54         \\
                              & 0.981          & 0.888          & 0.672          & 0.628          & 512        & 4      & 0.731      & 24         \\
                              & 0.990          & 0.870          & 0.786          & 0.767          & 256        & 5      & 0.704      & 35         \\
Delta-Exposure + DN      &                &                &                &                &            &        &            &            \\
                              & 0.990          & 0.869          & 0.787          & 0.764          & 512        & 4      & 0.525      & 21         \\
                              & 0.989          & 0.874          & 0.751          & 0.733          & 512        & 4      & 0.747      & 22         \\
Latency                       &                &                &                &                &            &        &            &            \\
                              & 0.982          & 0.929          & 0.736          & 0.691          & 128        & 6      & 0.246      & 10         \\
                              & 0.978          & 0.905          & 0.732          & 0.692          & 256        & 3      & 0.010      & 4          \\
{\ul Latency + DN}       &                &                &                &                &            &        &            &            \\
                              & \textbf{0.998} & 0.969          & \textbf{0.945} & \textbf{0.942} & 256        & 5      & 0.239      & 4          \\
Step-Forward-Direct           &                &                &                &                &            &        &            &            \\
                              & 0.986          & 0.877          & 0.733          & 0.717          & 512        & 6      & 0.102      & 9          \\
Step-Forward-Direct + DN &                &                &                &                &            &        &            &            \\
                              & 0.989          & 0.940          & 0.802          & 0.780          & 512        & 6      & 0.246      & 37         \\
                              & 0.990          & 0.918          & 0.836          & 0.827          & 256        & 5      & 0.025      & 33         \\
ANN                           &                &                &                &                &            &        &            &            \\
                              & 0.984          & 0.929          & 0.823          & 0.800          & 128        & 2      & \textbf{-} & \textbf{-} \\
                              & 0.977          & 0.958          & 0.778          & 0.694          & 256        & 3      & -          & -          \\
ANN + DN                 &                &                &                &                &            &        &            &            \\
                              & 0.994          & 0.976          & 0.922          & 0.914          & 512        & 3      & \textbf{-} & \textbf{-} \\
                              & 0.993          & \textbf{0.977} & 0.902          & 0.887          & 512        & 5      & \textbf{-} & \textbf{-} \\
                              & 0.994          & 0.976          & 0.921          & 0.914          & 512        & 3      & -          & - \\
                              \bottomrule
\end{tabular}
\footnotetext{We show the trials that performed best in at least one metric per encoding method. Best scores are in bold. Abbreviations: Divisive Normalisation `DN'; Artificial Neural Network `ANN'; Area Under the Receiver Operator Curve `AUROC', and Area Under the Precision Recall Curve `AUPRC'. `Num Hidden' refers to the number of neurons in each hidden layer, and `Num Layers' refers to the number of hidden layers. The best overall encoding method is underlined. Each entry scored the highest in at least one performance measure, hence the variable number of entries per encoding method.}
\end{sidewaystable}
\FloatBarrier
\subsection*{Effects of divisive normalisation on synthetic HERA dataset}
We took the best-performing trials as described in the methods section for each method and performed ten repeat trials to verify the performance of each.
Table \ref{tab:hera-trials} contains the results of these trials.
As in hyper-parameter optimisation, latency encoding with divisive normalisation performs strongest, followed by ANN networks with divisive normalisation.

\begin{table}[!htbp]
\centering
\caption{The final results show the performance of each encoding method using the final hyper-parameters on the HERA dataset with first-order LiF neurons.}
\label{tab:hera-trials}
\begin{tabularx}{\linewidth}{@{}ccccccccc@{}}
\toprule
Encoding Method               & \multicolumn{2}{c}{Accuracy} & \multicolumn{2}{c}{AUROC} & \multicolumn{2}{c}{AUPRC} & \multicolumn{2}{c}{F1} \\ \midrule
Delta Exposure                & 0.984             & 0.006    & 0.869            & 0.004  & 0.699            & 0.051  & 0.671          & 0.065          \\
Delta Exposure + DN      & 0.986             & 0.002    & 0.716            & 0.022  & 0.867            & 0.002  & 0.692          & 0.025          \\
Latency                       & 0.974             & 0.004    & 0.570            & 0.029  & 0.816            & 0.005  & 0.541          & 0.039          \\
{\ul{Latency + DN}}    & \textbf{0.996}    & 0.002    & 0.968            & 0.001  & 0.914            & 0.034  & \textbf{0.907} & 0.042          \\
Step-Forward-Direct           & 0.709             & 0.018    & \textbf{0.984}   & 0.001  & 0.878            & 0.004  & 0.687          & 0.023          \\
Step-Forward-Direct + DN & 0.990             & 0.002    & 0.823            & 0.037  & 0.936            & 0.001  & 0.805          & 0.046          \\
ANN                           & 0.879             & 0.067    & 0.645            & 0.079  & 0.894            & 0.031  & 0.440          & 0.155          \\
ANN + DN                 & 0.993             & 0.001    & 0.901            & 0.016  & \textbf{0.976}   & 0.001  & 0.886          & 0.023          \\ \bottomrule
\end{tabularx}
\footnotetext{Each metric is listed as mean and standard deviation. Ten trials were completed for each encoding method. Abbreviations: Divisive Normalisation `DN'; Artificial Neural Network `ANN'; Area Under the Receiver Operator Curve `AUROC', and Area Under the Precision Recall Curve `AUPRC'. The best scores are bolded. Latency encoding with divisive normalisation offers superior performance in two metrics. Divisive normalisation significantly improves accuracy, AUPRC, and F1 scores for all methods while degrading AUROC scores slightly for the delta-exposure and step-forward-direct encodings.}
\end{table}

\begin{figure}[!htbp]
    \centering
    \begin{subfigure}{0.45\columnwidth}
        \centering
        \includegraphics[height=1.5in, keepaspectratio]{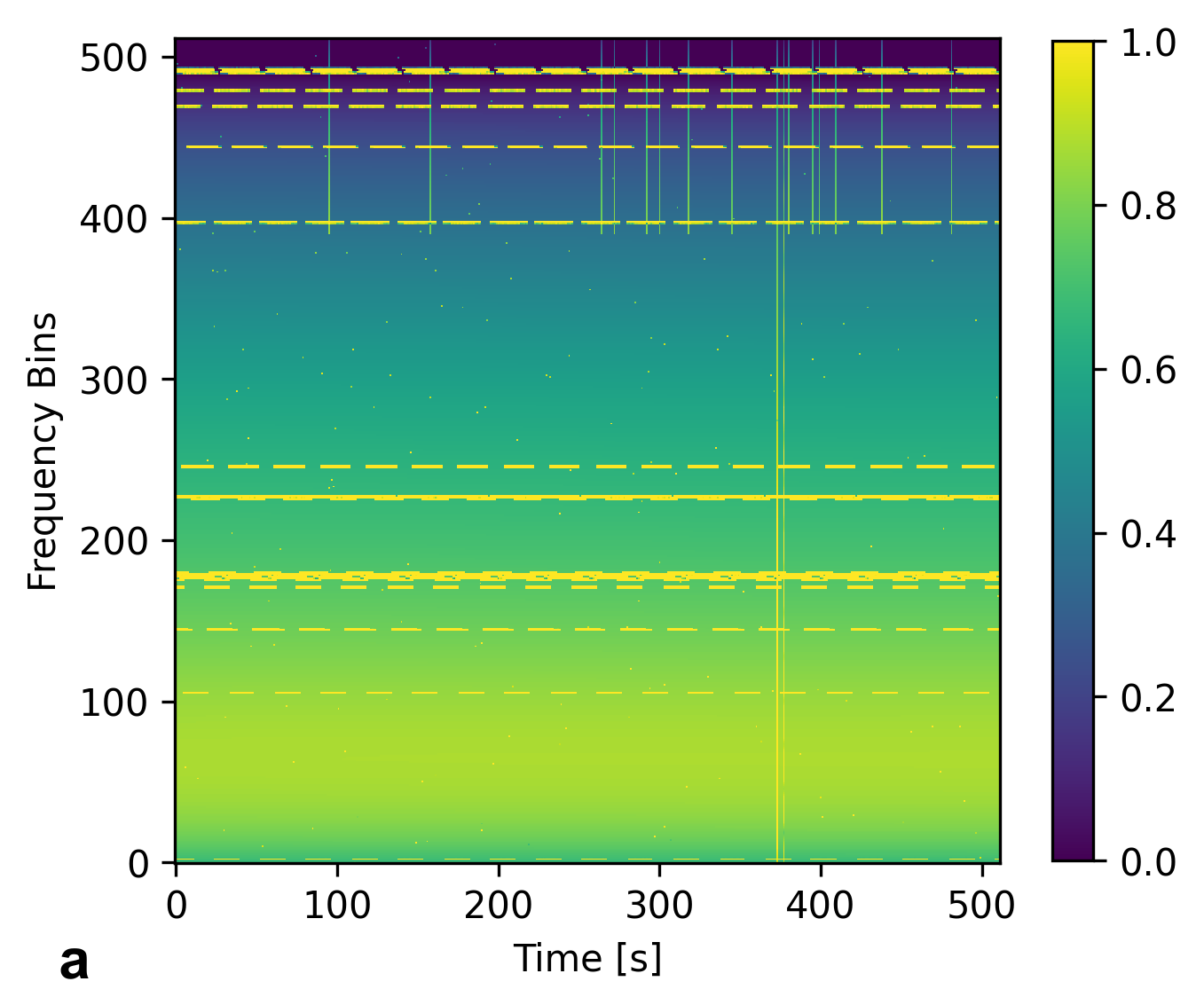}
        \label{fig:res:hera:orig}
    \end{subfigure}
    \begin{subfigure}{0.45\columnwidth}
        \centering
        \includegraphics[height=1.5in, keepaspectratio]{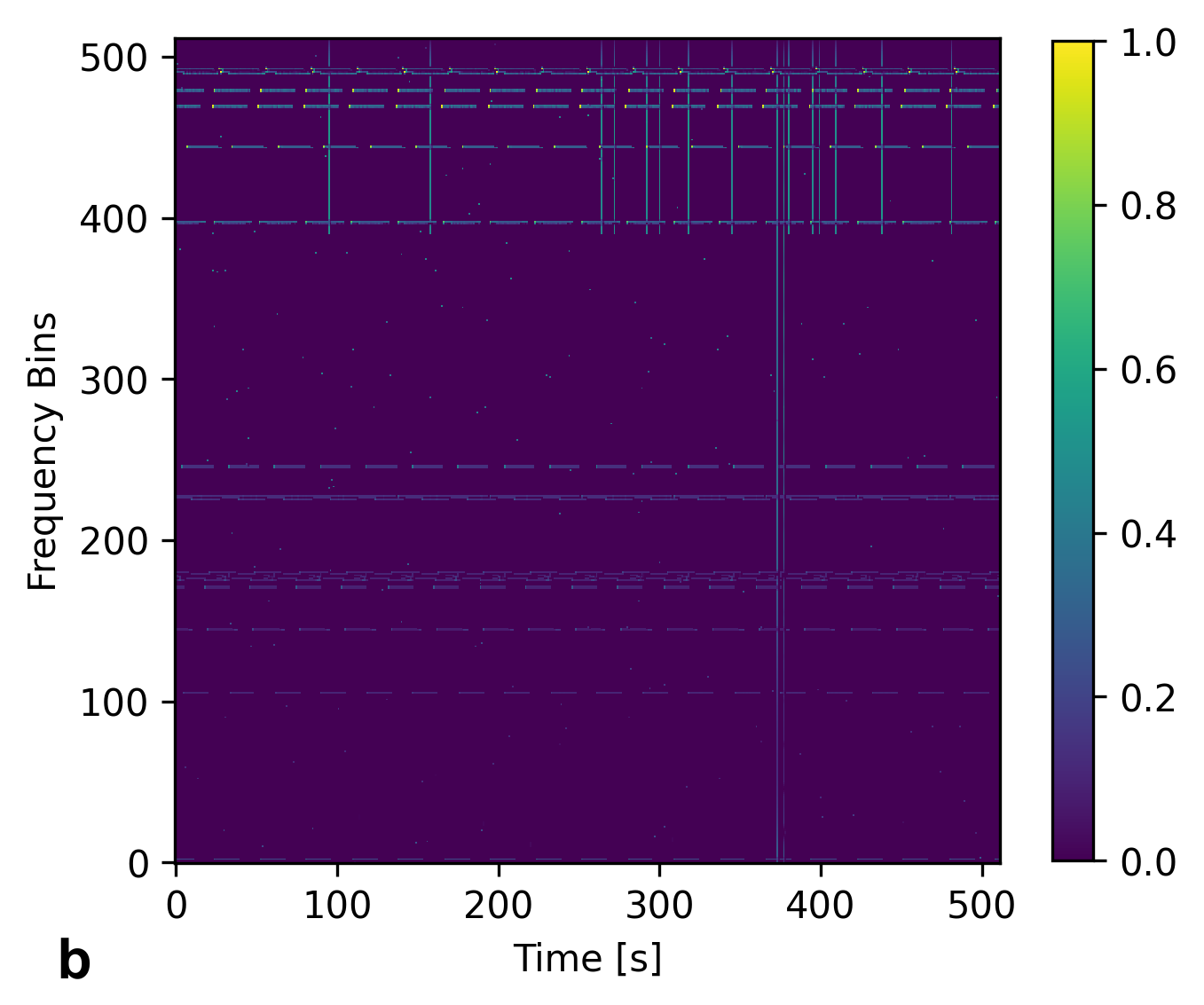}
        \label{fig:res:hera:divnorm}
    \end{subfigure}
    \begin{subfigure}{0.45\columnwidth}
        \centering
        \includegraphics[height=1.5in, keepaspectratio]{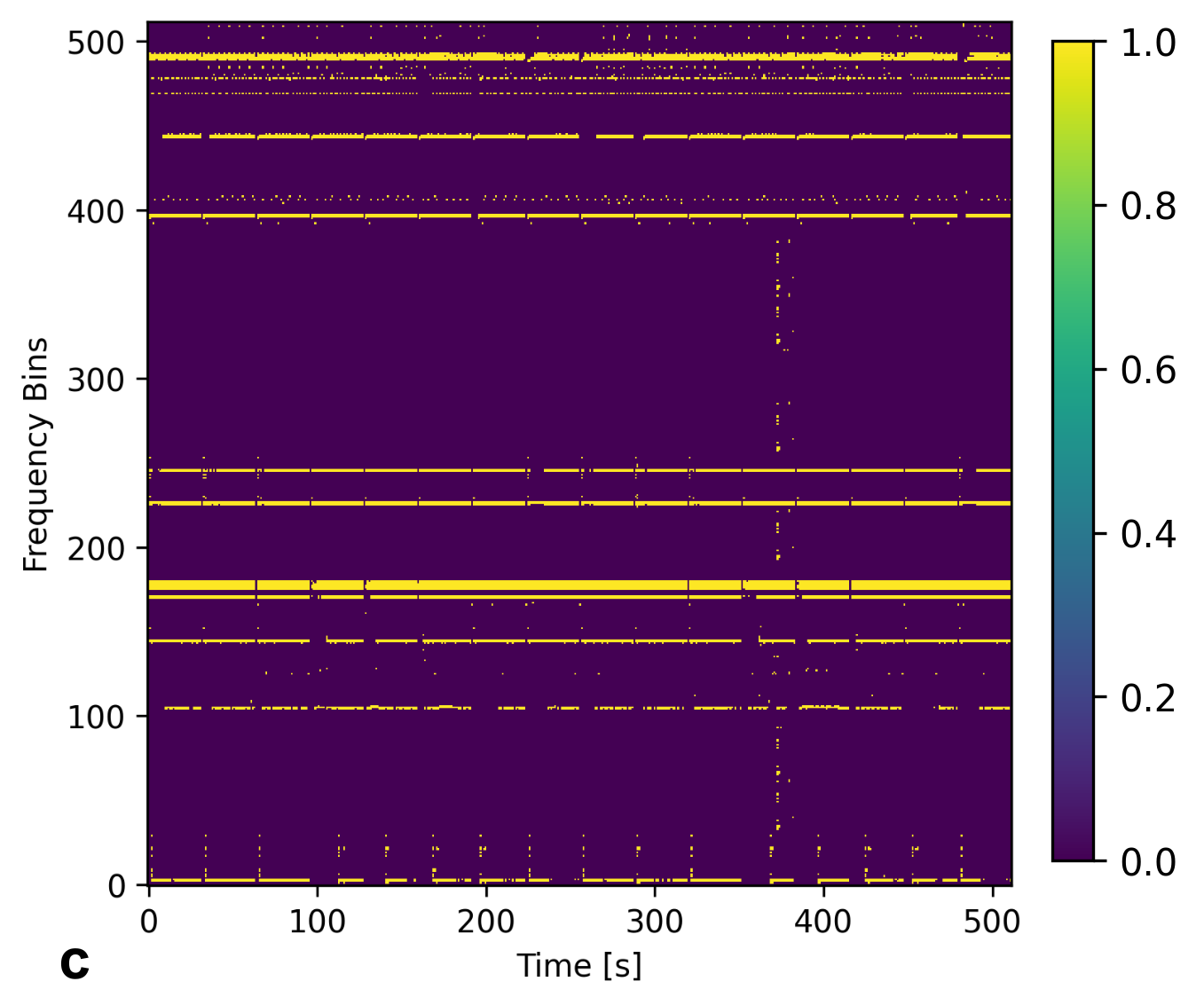}
        \label{fig:res:hera:inf-orig}
    \end{subfigure}
    \begin{subfigure}{0.45\columnwidth}
        \centering
        \includegraphics[height=1.5in, keepaspectratio]{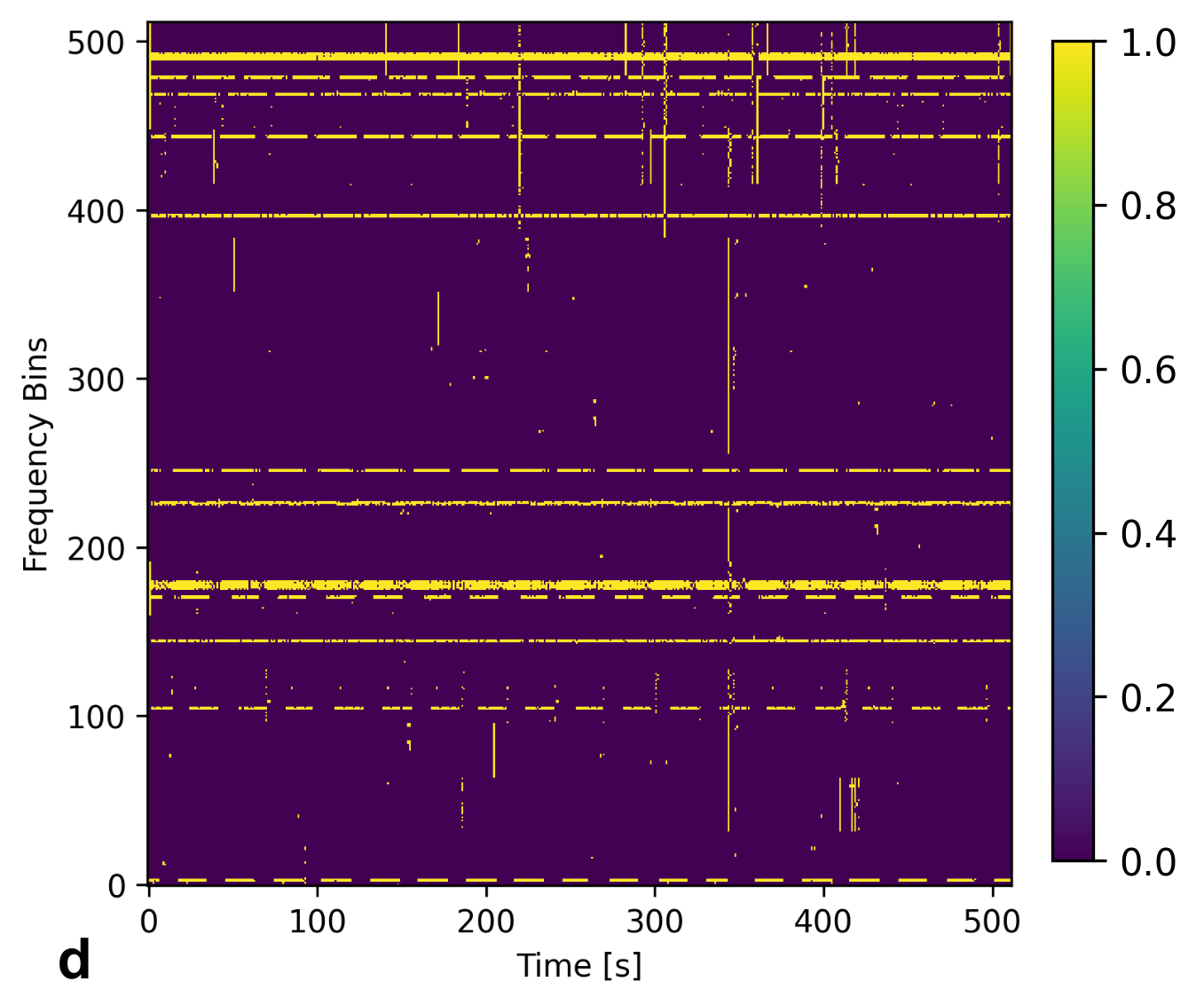}
        \label{fig:res:hera:inf-divnorm}
    \end{subfigure}
    \begin{subfigure}{0.45\columnwidth}
        \centering
        \includegraphics[height=1.5in, keepaspectratio]{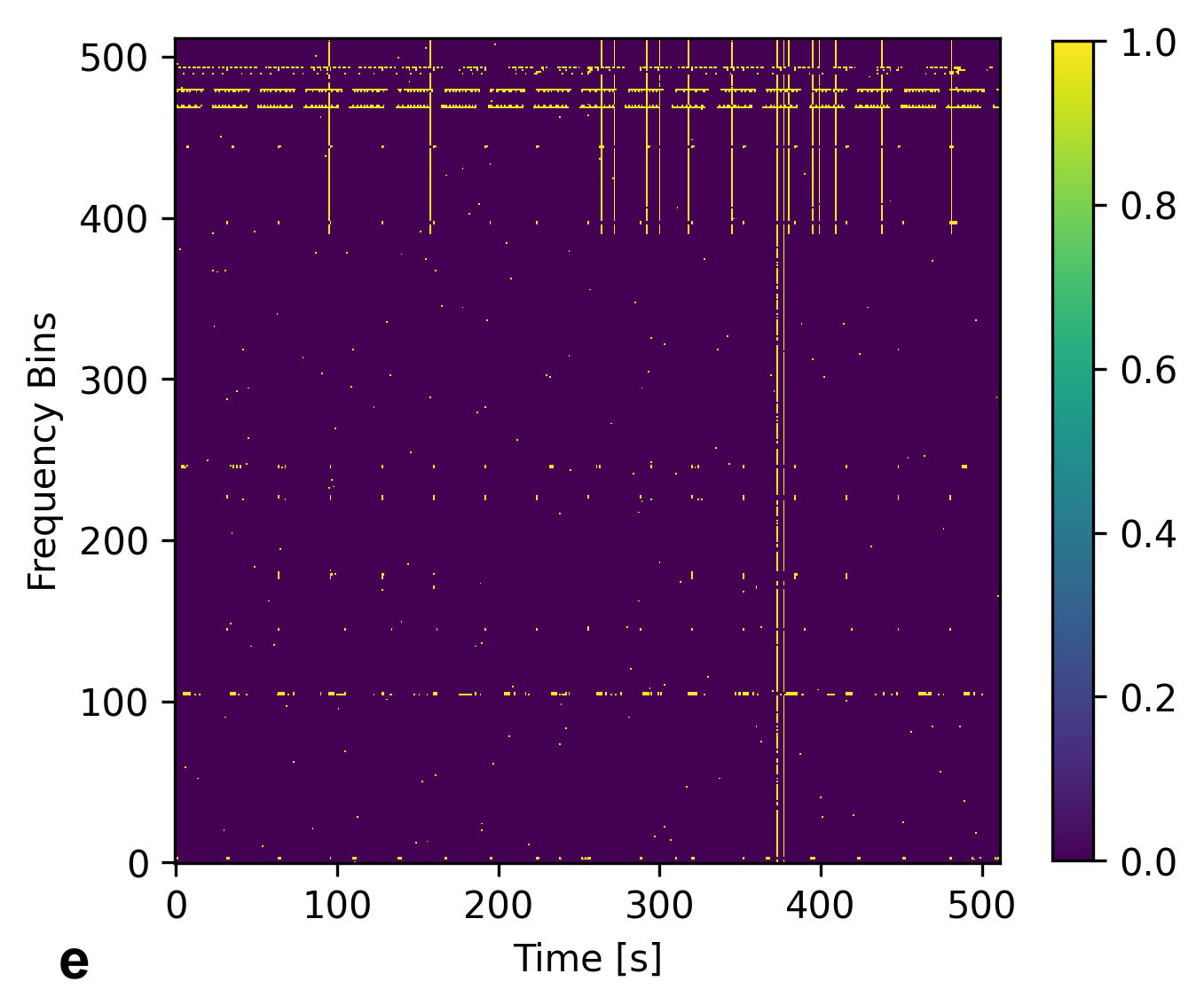}
        \label{fig:res:hera:residual}
    \end{subfigure}
    \begin{subfigure}{0.45\columnwidth}
        \centering
        \includegraphics[height=1.5in, keepaspectratio]{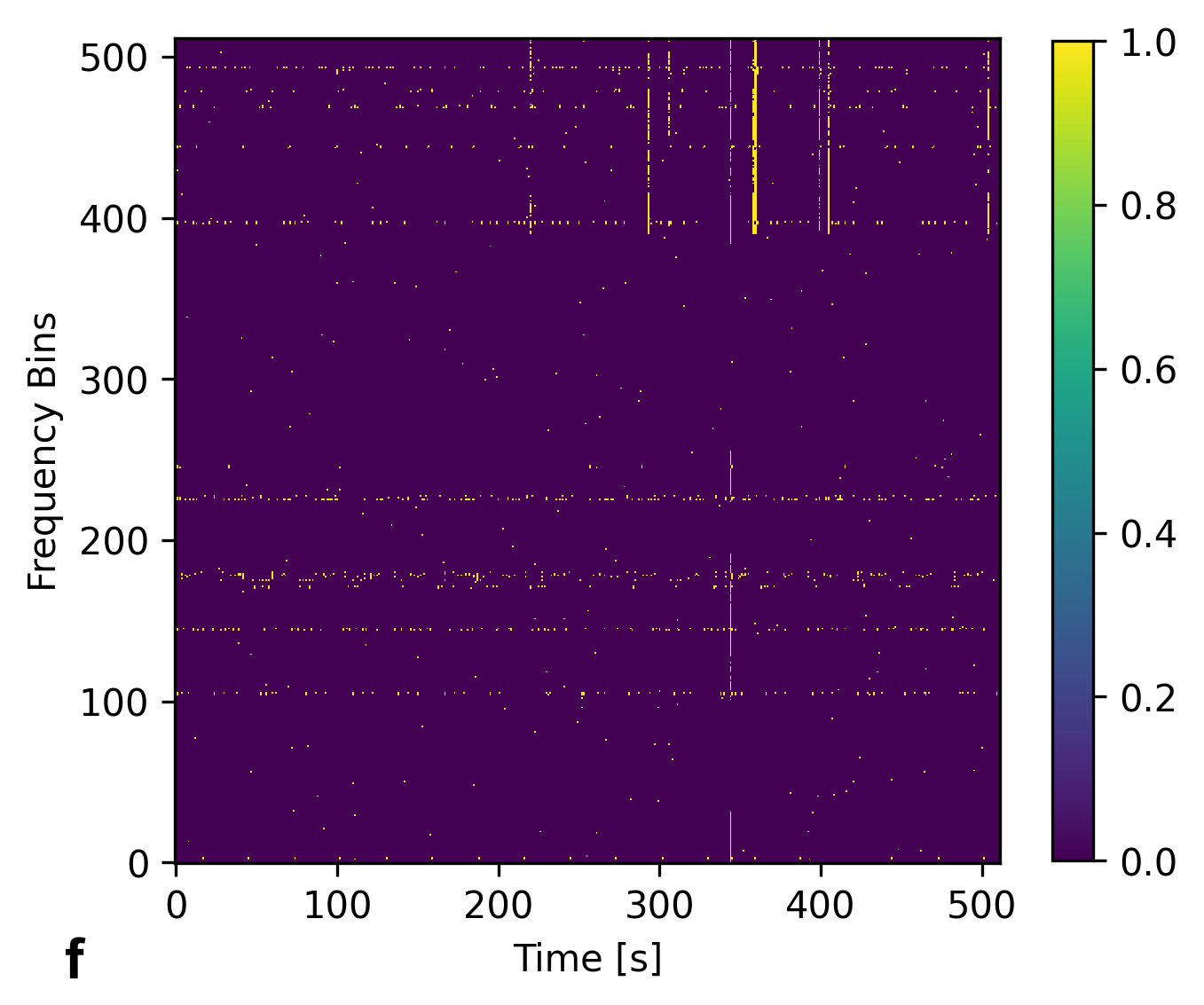}
        \label{fig:res:hera:divnorm:residual}
    \end{subfigure}
    \begin{subfigure}{0.45\columnwidth}
        \centering
        \includegraphics[height=1.5in, keepaspectratio]{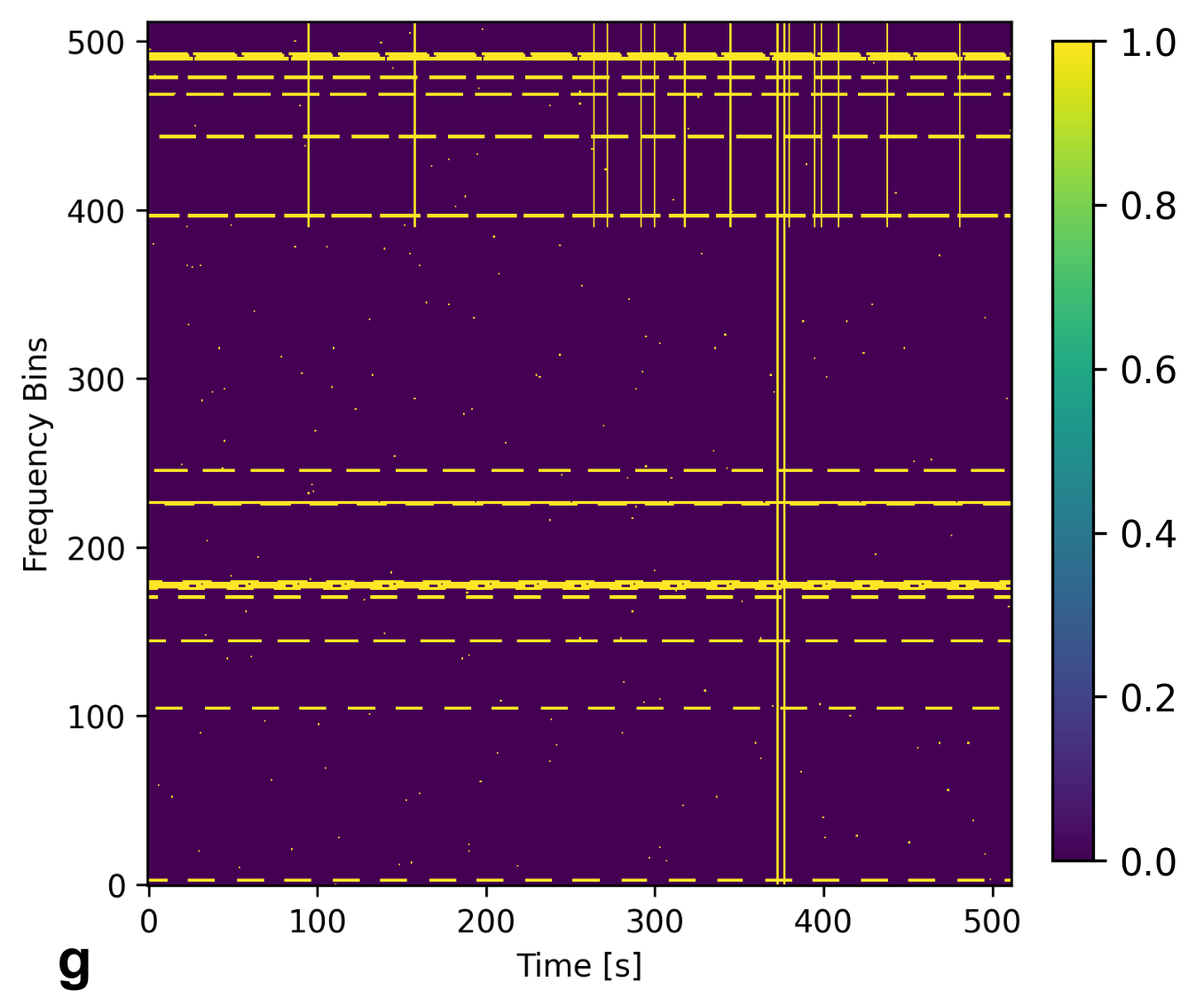}
        \label{fig:res:hera:mask}
    \end{subfigure}
    \caption{Impact of Divisive Normalisation on Radio Frequency Inference (RFI) Detection in HERA Spectrograms. Sub-panels depict: (a) original spectrogram; (b) spectrogram after divisive normalisation, with background gradients reduced; (c) latency-based inference on the original spectrogram; (d) latency-based inference on the normalised spectrogram showing clearer RFI features; (e) residual between original inference and mask; (f) residual between normalised inference and mask, and (g) ground-truth RFI mask. Divisive normalisation significantly reduces background noise while preserving key RFI features, improving inference performance of the SNN.}
\label{fig:example:hera}
\end{figure}

Figure \ref{fig:example:hera} presents an example spectrogram original and divisible normalised, latency encoded inference, and the original ground-truth RFI mask.
This figure clearly shows that divisive normalisation removes almost all of the background `gradient' information representing the simulated astronomical background from the dataset, resulting in significantly clearer inputs and outputs from the SNN.
Interestingly, we find using a more complex second-order LiF neuron model degrades performance significantly on this dataset and provide these results in Supplementary Table 1.
On average, divisive normalisation improved accuracy by 0.102, AUROC by 0.029, AUPRC by 0.108 and F1-Score by 0.168 across each spike-encoding method.
We find latency encoding with divisive normalisation to be the strongest performing encoding method with an AUROC of 0.968, AUPRC of 0.914 and F1-Score of 0.907 to be comparable with state-of-the-art traditional algorithms, ANN methods and ANN2SNN conversions on this dataset.
Table \ref{tab:res:hera:compare} contains a comparison between this work and the current state of the art algorithmic, ANN and SNN performances on a HERA-derived dataset.
We see that despite the drastically reduced model size, our directly trained BPTT + DN latency encoded model performs competitively against existing methods.
\begin{table}[!htbp]
\centering
\caption{Comparison to other algorithmic, ANN and SNN baselines on the HERA dataset.}
\label{tab:res:hera:compare}
\begin{tabular}{@{}cccccc@{}}
\toprule
Work                                             & Model                                                          & Method    & AUROC & AUPRC & F1    \\ \midrule
\cite{mesarcik_learning_2022} & AOFlagger                                                      & Algorithm & 0.974 & 0.880 & 0.873 \\
\cite{vanzyl_remove_2024}     & Auto-Encoder                                                   & ANN       & 0.994 & 0.959 & 0.945 \\
\cite{pritchard_rfi_2024}     & \begin{tabular}[c]{@{}l@{}}Auto-Encoder\end{tabular} & ANN2SNN       & 0.944 & 0.920 & 0.953 \\
This work                                        & BPTT + DN                                                      & SNN       & 0.968 & 0.914 & 0.907 \\ \bottomrule
\end{tabular}
\footnotetext{Abbreviations: Divisive Normalisation `DN'; Spiking Neural Network `SNN'; Artificial Neural Network `ANN'; Backpropagation Through Time `BPTT'; Area Under the Receiver Operator Curve `AUROC', and Area Under the Precision Recall Curve `AUPRC'. We see that directly trained SNNs perform comparably to the state of the art, which is noteworthy considering their relatively small size.}
\end{table}
\FloatBarrier\clearpage
\subsection*{Hyper-parameter tuning on real LOFAR dataset}
We optimised hyper-parameters on a real LOFAR-derived dataset with the same methodology as the synthetic HERA dataset.
The methods section contains full details of this realistic dataset.
Being based in population-dense Europe, the RFI environment of this instrument is significantly more challenging than in the synthetic dataset. 
Table \ref{tab:lofar:optuna} contains results from our hyper-parameter search on the real LOFAR dataset.
We find performance across the board degraded significantly compared to the synthetic case where the ANN with and without divisive normalisation performs best in a single trial.
The optimiser favours slightly shallower networks with faster decay rates, wider hidden layers and longer exposure times across all encoding methods.
\begin{sidewaystable}[!htbp]
\caption{Hyper-parameter search using Optuna multi-variate optimisation for the real LOFAR dataset.}
\label{tab:lofar:optuna}
\begin{tabular}{ccccccccc}
\hline
Encoding Method               & \multicolumn{1}{c}{Accuracy} & \multicolumn{1}{c}{AUROC} & \multicolumn{1}{c}{AUPRC} & \multicolumn{1}{c}{F1-Score} & \multicolumn{1}{c}{\makecell{Num\\Hidden}} & \multicolumn{1}{c}{\makecell{Num\\Layers}} & \multicolumn{1}{c}{Beta}       & \multicolumn{1}{c}{Exposure}   \\ \hline
Delta-Exposure                & \multicolumn{1}{c}{}         & \multicolumn{1}{c}{}      & \multicolumn{1}{c}{}      & \multicolumn{1}{c}{}         & \multicolumn{1}{c}{}           & \multicolumn{1}{c}{}       & \multicolumn{1}{c}{}           & \multicolumn{1}{c}{}           \\
                              & 0.975                        & 0.568                     & 0.163                     & 0.127                        & 512                            & 2                          & 0.736                          & 57                             \\
                              & 0.988                        & 0.539                     & 0.158                     & 0.146                        & 128                            & 2                          & 0.162                          & 18                             \\
                              & 0.986                        & 0.561                     & 0.173                     & 0.158                        & 512                            & 5                          & 0.374                          & 62                             \\
Delta-Exposure + DN      & \multicolumn{1}{c}{}         & \multicolumn{1}{c}{}      & \multicolumn{1}{c}{}      & \multicolumn{1}{c}{}         & \multicolumn{1}{c}{}           & \multicolumn{1}{c}{}       & \multicolumn{1}{c}{}           & \multicolumn{1}{c}{}           \\
                              & 0.986                        & 0.514                     & 0.100                     & 0.093                        & 256                            & 5                          & 0.840                          & 11                             \\
                              & 0.970                        & 0.563                     & 0.152                     & 0.111                        & 512                            & 2                          & 0.735                          & 55                             \\
\multicolumn{1}{l}{}          & 0.974                        & 0.551                     & 0.141                     & 0.114                        & 512                            & 3                          & 0.742                          & 23                             \\
Latency                       & \multicolumn{1}{c}{}         & \multicolumn{1}{c}{}      & \multicolumn{1}{c}{}      & \multicolumn{1}{c}{}         & \multicolumn{1}{c}{}           & \multicolumn{1}{c}{}       & \multicolumn{1}{c}{}           & \multicolumn{1}{c}{}           \\
                              & 0.988                        & 0.568                     & 0.248                     & 0.211                        & 512                            & 2                          & 0.149                          & 61                             \\
                              & 0.977                        & 0.610                     & 0.226                     & 0.175                        & 512                            & 2                          & 0.592                          & 52                             \\
Latency + DN             & \multicolumn{1}{c}{}         & \multicolumn{1}{c}{}      & \multicolumn{1}{c}{}      & \multicolumn{1}{c}{}         & \multicolumn{1}{c}{}           & \multicolumn{1}{c}{}       & \multicolumn{1}{c}{}           & \multicolumn{1}{c}{}           \\
                              & 0.981                        & 0.617                     & 0.240                     & 0.198                        & 512                            & 3                          & 0.418                          & 63                             \\
\multicolumn{1}{l}{}          & 0.976                        & 0.628                     & 0.244                     & 0.184                        & 256                            & 2                          & 0.675                          & 62                             \\
\multicolumn{1}{l}{}          & 0.989                        & 0.550                     & 0.218                     & 0.188                        & 512                            & 5                          & 0.109                          & 23                             \\
Step-Forward-Direct           & \multicolumn{1}{c}{}         & \multicolumn{1}{c}{}      & \multicolumn{1}{c}{}      & \multicolumn{1}{c}{}         & \multicolumn{1}{c}{}           & \multicolumn{1}{c}{}       & \multicolumn{1}{c}{}           & \multicolumn{1}{c}{}           \\
                              & 0.970                        & 0.636                     & 0.247                     & 0.171                        & 512                            & 2                          & 0.441                          & 29                             \\
\multicolumn{1}{l}{}          & 0.986                        & 0.624                     & 0.283                     & 0.246                        & 512                            & 2                          & 0.512                          & 24                             \\
Step-Forward-Direct + DN & \multicolumn{1}{c}{}         & \multicolumn{1}{c}{}      & \multicolumn{1}{c}{}      & \multicolumn{1}{c}{}         & \multicolumn{1}{c}{}           & \multicolumn{1}{c}{}       & \multicolumn{1}{c}{}           & \multicolumn{1}{c}{}           \\
                              & 0.971                        & 0.628                     & 0.238                     & 0.167                        & 512                            & 3                          & 0.263                          & 42                             \\
\multicolumn{1}{l}{}          & 0.968                        & 0.633                     & 0.241                     & 0.162                        & 512                            & 4                          & 0.373                          & 30                             \\
                              & 0.973                        & 0.583                     & 0.183                     & 0.140                        & 512                            & 6                          & 0.078                          & 62                             \\
{\ul ANN}                     & \multicolumn{1}{c}{}         & \multicolumn{1}{c}{}      & \multicolumn{1}{c}{}      & \multicolumn{1}{c}{}         & \multicolumn{1}{c}{}           & \multicolumn{1}{c}{}       & \multicolumn{1}{c}{}           & \multicolumn{1}{c}{}           \\
                              & \textbf{0.992}               & 0.485                     & \textbf{0.519}            & 0.047                        & 512                            & 6                          & \multicolumn{1}{c}{\textbf{-}} & \multicolumn{1}{c}{\textbf{-}} \\
                              & 0.977                        & 0.761                     & 0.450                     & \textbf{0.356}               & 512                            & 3                          & \multicolumn{1}{c}{-}          & \multicolumn{1}{c}{-}          \\
\multicolumn{1}{l}{}          & 0.976                        & \textbf{0.765}            & 0.451                     & 0.351                        & 512                            & 3                          &                                &                                \\
ANN + DN                 & \multicolumn{1}{c}{}         & \multicolumn{1}{c}{}      & \multicolumn{1}{c}{}      & \multicolumn{1}{c}{}         & \multicolumn{1}{c}{}           & \multicolumn{1}{c}{}       & \multicolumn{1}{c}{}           & \multicolumn{1}{c}{}           \\
                              & 0.674                        & 0.673                     & 0.403                     & 0.114                        & 512                            & 4                          & \multicolumn{1}{c}{\textbf{-}} & \multicolumn{1}{c}{\textbf{-}} \\
                              & 0.719                        & 0.689                     & 0.389                     & 0.090                        & 256                            & 2                          & \multicolumn{1}{c}{\textbf{-}} & \multicolumn{1}{c}{\textbf{-}} \\
                              & \textbf{0.992}               & 0.485                     & \textbf{0.519}            & 0.047                        & 512                            & 6                          & \multicolumn{1}{c}{-}          & \multicolumn{1}{c}{-}          \\ \hline
\end{tabular}
\footnotetext{We show the trials that performed best in at least one metric per encoding method. The best scores are in bold. Abbreviations: Divisive Normalisation `DN'; Artificial Neural Network `ANN'; Area Under the Receiver Operator Curve `AUROC', and Area Under the Precision Recall Curve `AUPRC'. `Num Hidden' refers to the number of neurons in each hidden layer, and `Num Layers' refers to the number of hidden layers. The best overall encoding method is underlined. Each entry scored the highest in at least one performance measure, hence the variable number of entries per encoding method.}
\end{sidewaystable}
\FloatBarrier
\subsection*{Effects of divisive normalisation on real LOFAR dataset}
In line with our approach to the HERA dataset, we took the best-performing parameters for each method and performed repeat trials to verify performance.
Table \ref{tab:lofar-trials} contains the results of these trials.
We see that, on aggregate, our SNN-encoded methods perform better than the equivalently sized ANN methods.
The best-performing method overall is latency encoding; however, the significantly better AUPRC score from step-forward (direct) encoding with divisive normalisation is noteworthy.
\begin{table}[!htbp]
\centering
\caption{The final results show the performance of each encoding method using the final hyper-parameters on the LOFAR dataset with second-order LiF neurons.}
\label{tab:lofar-trials}
\begin{tabularx}{\linewidth}{@{}cllllllll@{}}
\toprule
Encoding Method               & \multicolumn{2}{c}{Accuracy} & \multicolumn{2}{c}{AUROC} & \multicolumn{2}{c}{AUPRC} & \multicolumn{2}{c}{F1} \\ \midrule
Delta Exposure                & 0.887             & 0.294    & 0.314            & 0.007  & 0.621            & 0.108  & 0.397          & 0.009          \\
Delta Exposure + DN      & 0.992             & 0.000    & 0.312            & 0.000  & 0.629            & 0.011  & 0.395          & 0.001          \\
{\ul{Latency}}                       & \textbf{0.992}             & 0.002    & 0.346            & 0.012  & 0.604            & 0.021  & \textbf{0.474}          & 0.024          \\
Latency + DN    & 0.989             & 0.003    & 0.335   & 0.005  & 0.475            & 0.007  & 0.438          & 0.005          \\
Step-Forward-Direct           & 0.989    & 0.006    & 0.351            & 0.008  & 0.497   & 0.015  & 0.469 & 0.012          \\
Step-Forward-Direct + DN & 0.992             & 0.000    & 0.311            & 0.001  & \textbf{0.693}            & 0.000  & 0.393          & 0.002          \\
ANN                           & 0.748             & 0.165    & 0.439            & 0.053  & 0.608            & 0.083  & 0.087          & 0.027          \\
ANN + DN                 & 0.862             & 0.160    & \textbf{0.473}            & 0.056  & 0.559            & 0.091  & 0.072          & 0.032          \\\bottomrule
\end{tabularx}
\footnotetext{Each metric is listed as mean and standard deviation. The best scores are bolded. Ten trials were completed for each encoding method. Abbreviations: Divisive Normalisation `DN'; Artificial Neural Network `ANN'; Area Under the Receiver Operator Curve `AUROC', and Area Under the Precision Recall Curve `AUPRC'.
Latency encoding without divisive normalisation offers the best overall performance in accuracy and F1 scores.
Step-Forward (direct) encoding with divisive normalisation offers significantly higher AUPRC performance than all other methods. An identically sized ANN does perform best in AUROC however.}
\end{table}

\begin{figure}[!htbp]
    \centering
    \begin{subfigure}{0.45\columnwidth}
        \centering
        \includegraphics[height=1.5in, keepaspectratio]{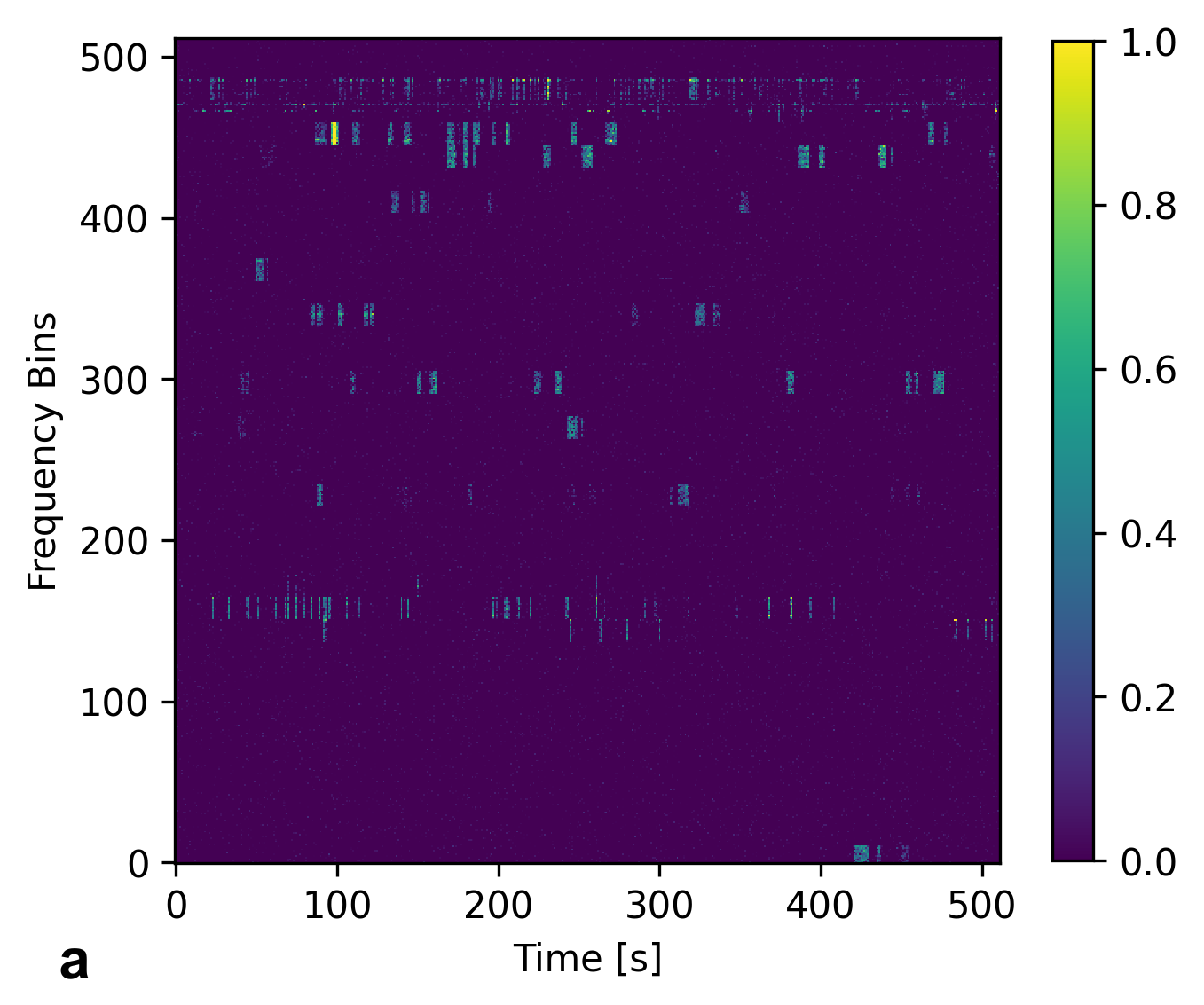}
        \label{fig:res:lofar:orig}
    \end{subfigure}
    \begin{subfigure}{0.45\columnwidth}
        \centering
        \includegraphics[height=1.5in, keepaspectratio]{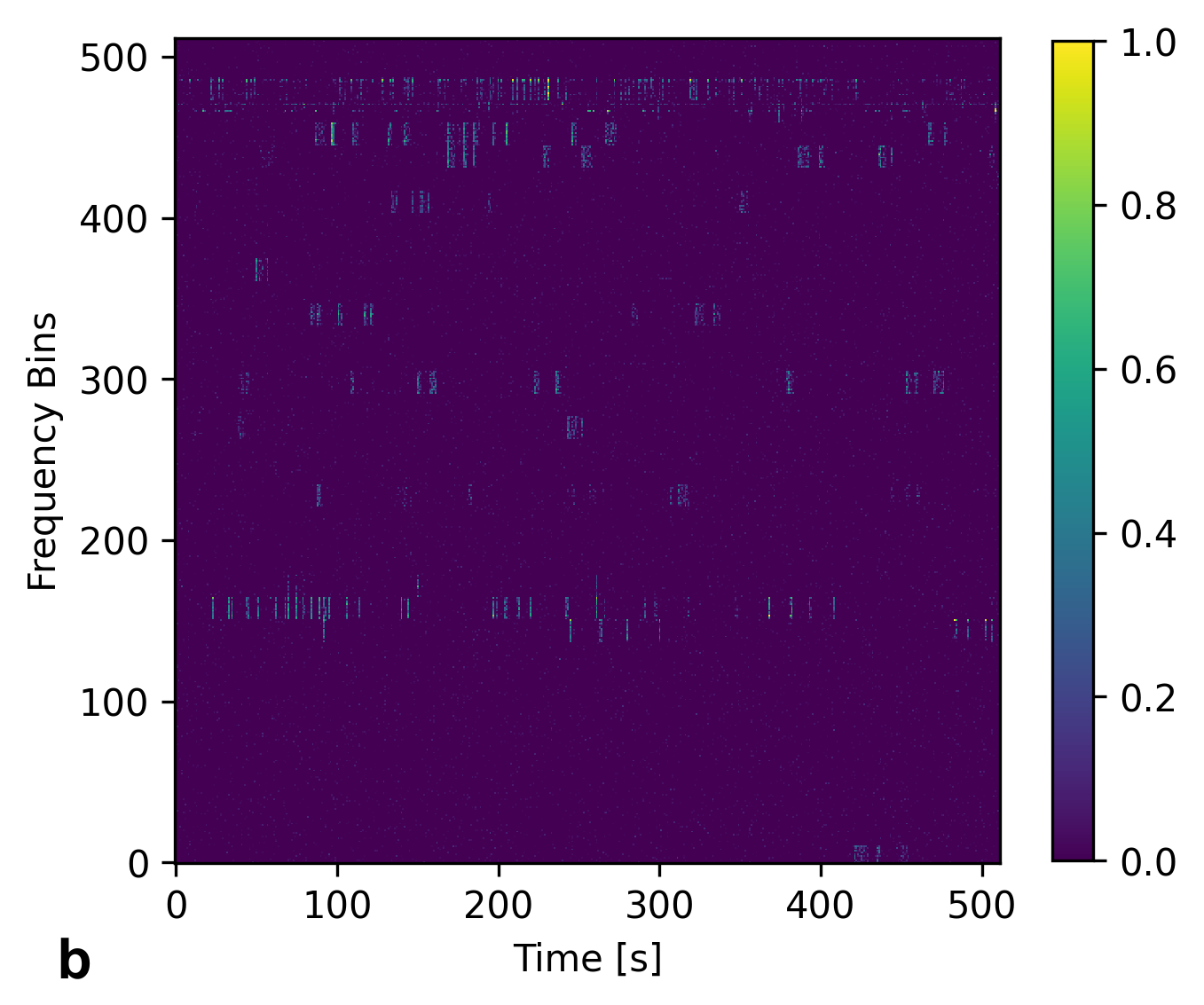}
        \label{fig:res:lofar:divnorm}
    \end{subfigure}
    \begin{subfigure}{0.45\columnwidth}
        \centering
        \includegraphics[height=1.5in, keepaspectratio]{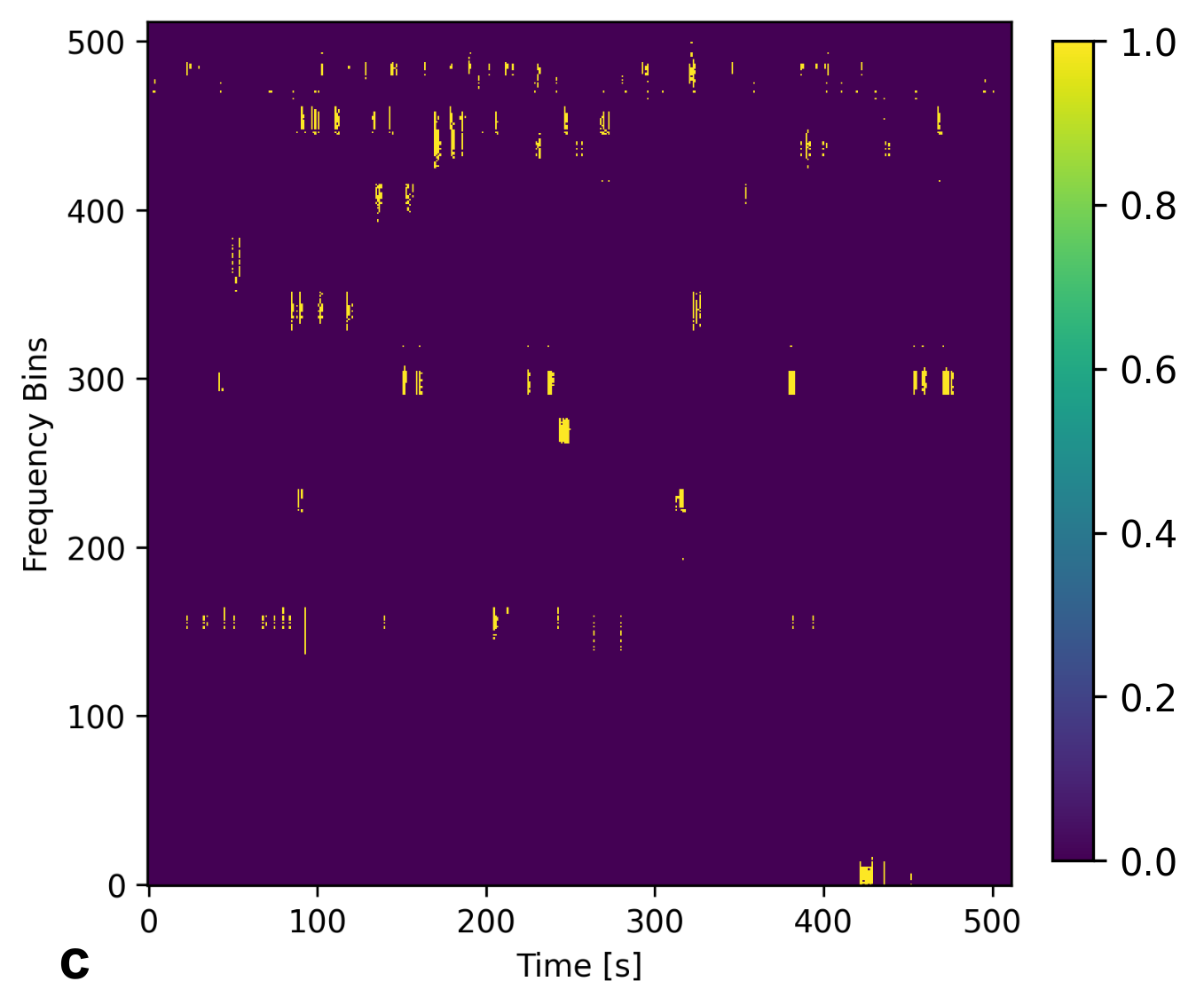}
        \label{fig:res:lofar:inf-orig}
    \end{subfigure}
    \begin{subfigure}{0.45\columnwidth}
        \centering
        \includegraphics[height=1.5in, keepaspectratio]{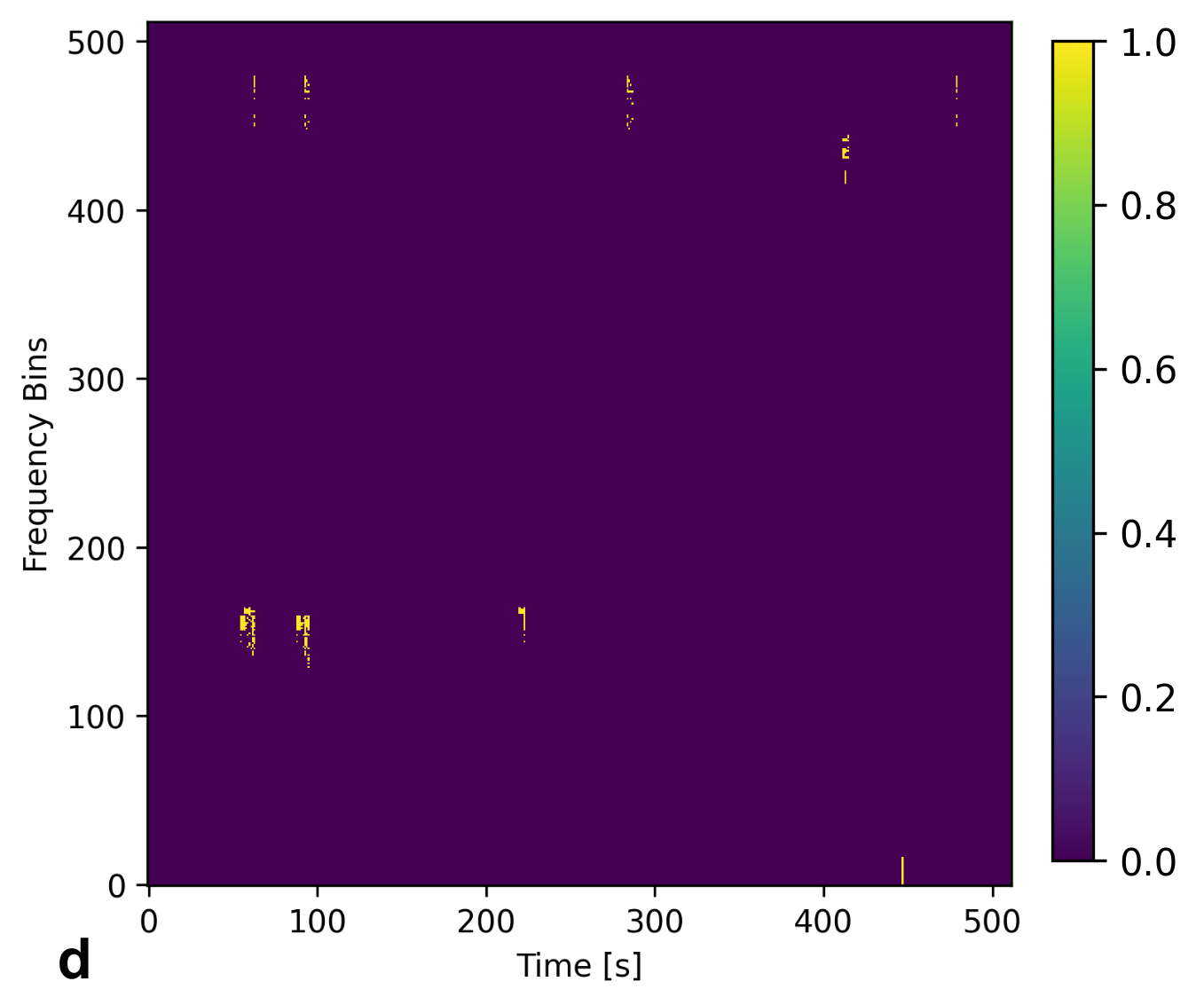}
        \label{fig:res:lofar:inf-divnorm}
    \end{subfigure}
    \begin{subfigure}{0.45\columnwidth}
        \centering
        \includegraphics[height=1.5in, keepaspectratio]{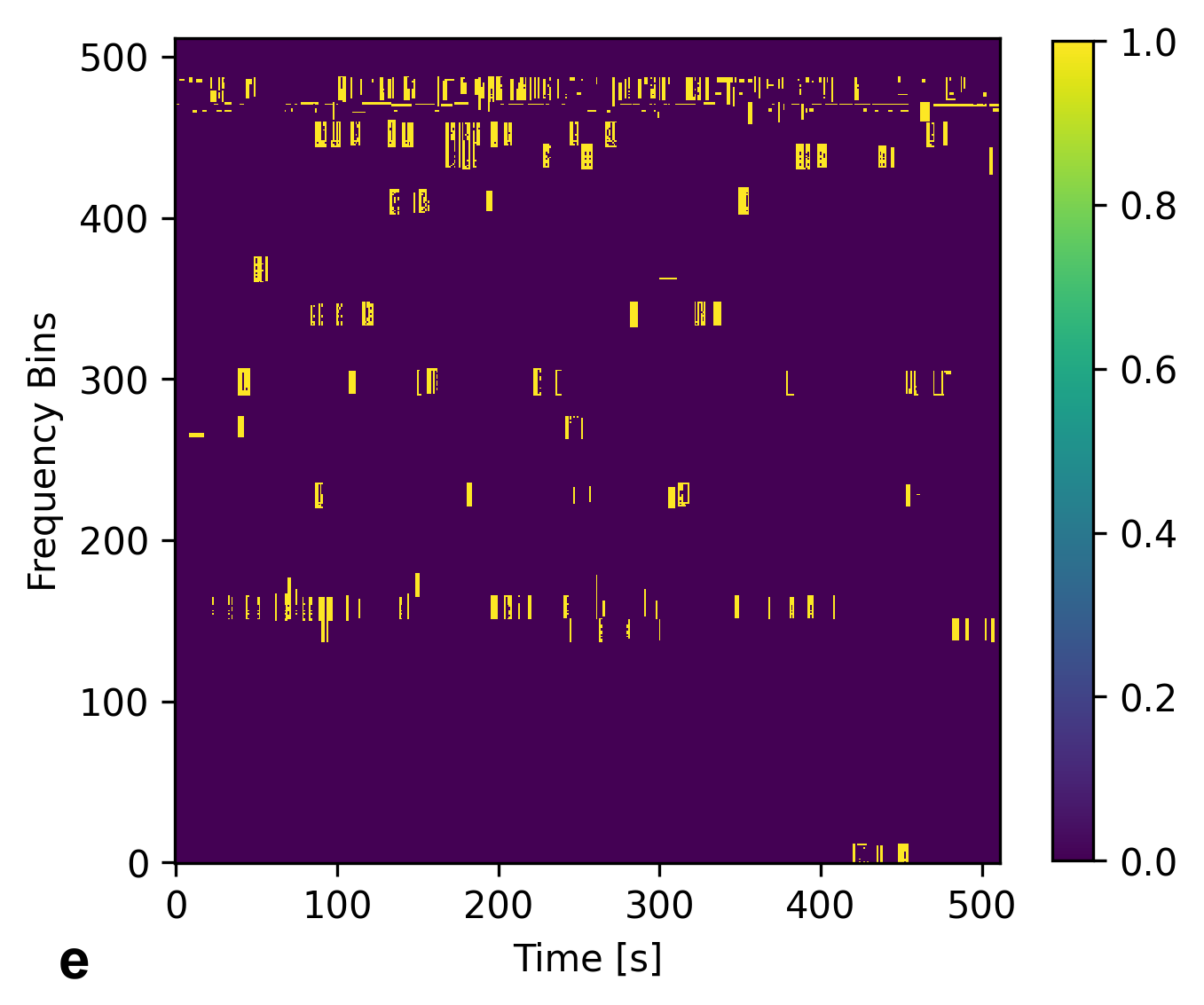}
        \label{fig:res:lofar:residual}
    \end{subfigure}
    \begin{subfigure}{0.45\columnwidth}
        \centering
        \includegraphics[height=1.5in, keepaspectratio]{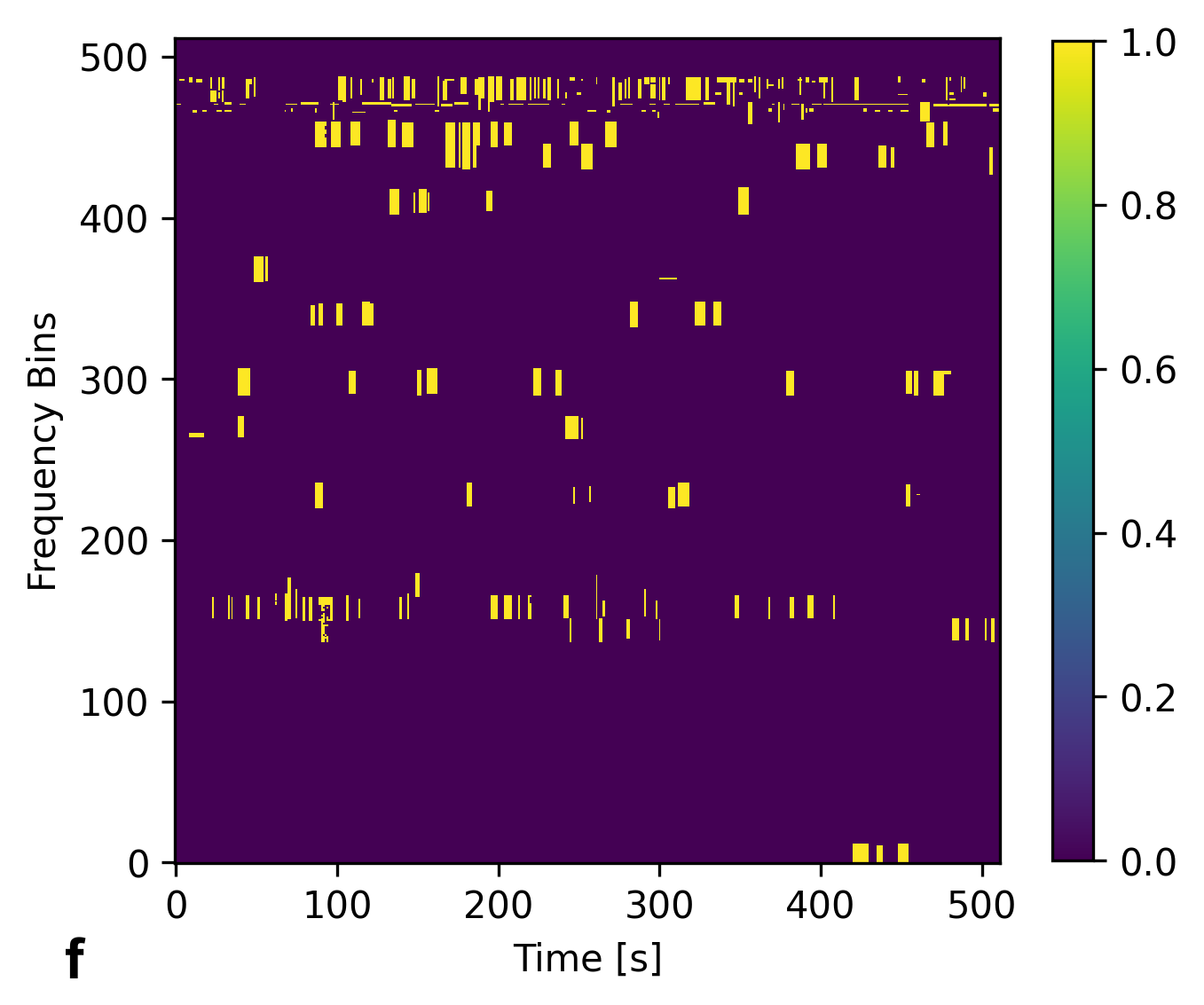}
        \label{fig:res:lofar:divnorm:residual}
    \end{subfigure}
    \begin{subfigure}{0.45\columnwidth}
        \centering
        \includegraphics[height=1.5in, keepaspectratio]{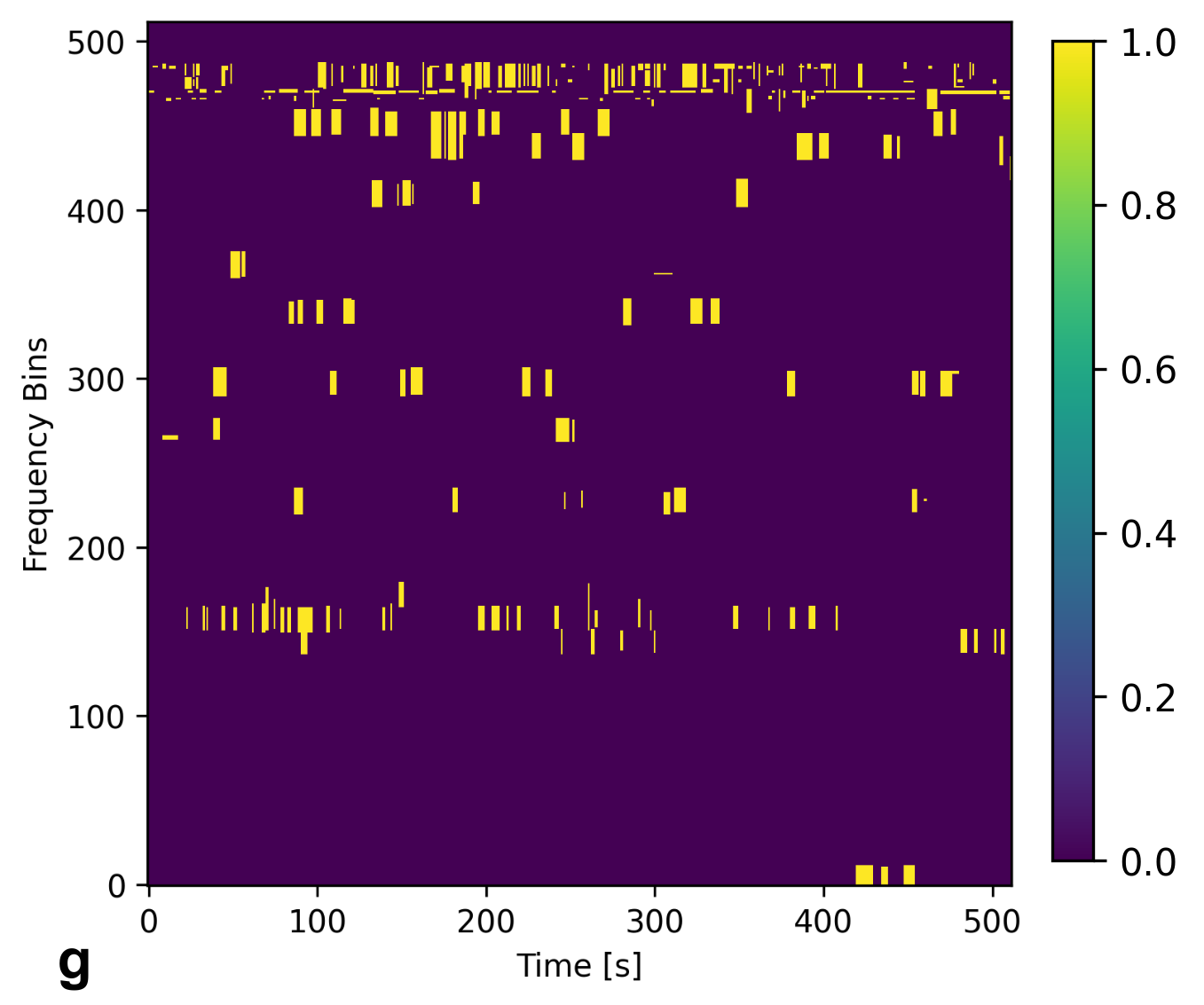}
        \label{fig:res:lofar:mask}
    \end{subfigure}
    \caption{Impact of Divisive Normalisation on Radio Frequency Interferrence (RFI) Detection in LOFAR Spectrograms. Sub-panels depict: (a) original spectrogram; (b) spectrogram after divisive normalisation; (c) latency-based inference on the original spectrogram; (d) latency-based inference on the normalised spectrogram showing a significantly more conservative flagging approach; (e) residual between original inference and mask; (f) residual between normalised inference and mask, and (g) expert-labelled RFI mask. While divisive normalisation improves clarity, the SNN output still exhibits significant noise compared to expert labelling. This is most apparent in the residual plots where in both cases key features are missing from inference.}
\label{fig:example:lofar}
\end{figure}

In Figure \ref{fig:example:lofar}, the SNN outputs are noisy with or without divisive normalisation, although the inference trained on normalised data shows the RFI features more prominently.
We also note the observation of banding in frequency and time dimensions near patch boundaries in some trials, which is the result of biases learned by these network instances and a fuzziness in the decoding process, which is the result of attempting learning on a significantly more complex dataset.
We discuss this phenomena and other preprocessing techniques, which have been key to achieving competitive performance in the Methods section.
Moreover, we provide significantly poorer results using the simpler first-order LiF neuron model in Supplementary Table 2.
While promising, the noisy output of the SNN is likely contributing to a degraded performance.
On average, divisive normalisation increased accuracy by 0.035, decreased AUROC by 0.017, increased AUPRC by 0.025 but decreased F1-Scores by 0.038 across all spike-encoding methods.
We find Latency encoding to give the strongest performance with an AUROC of 0.346, AUPRC of 0.604, and F1-Score of 0.474.
These results are compelling for such relatively lightweight architectures, however there exists a gap to the state-of-the-art.
Table \ref{tab:res:lofar:compare} compares the results of our best-performing directly trained SNN (Step-Forward with direct encoding) with existing algorithmic, ANN, and SNN baselines.
We see that there is a sizeable gap to the state of the art, indicating potential for this time-series segmentation approach to apply to real-world data, however some additional developments are needed to close the gap to the state-of-the-art.
Notably however, these results exceed prior ANN2SNN conversion-based results.
\begin{table}[!htbp]
\centering
\caption{Comparison to other algorithmic, ANN, and SNN baselines on the LOFAR dataset.}
\label{tab:res:lofar:compare}
\begin{tabular}{@{}cccccc@{}}
\toprule
Work                          & Model                                                          & Method    & AUROC & AUPRC & F1    \\ \midrule
\cite{mesarcik_learning_2022} & AOFlagger                                                      & Algorithm & 0.788 & 0.572 & 0.570 \\
\cite{dutoit_comparison_2024} & Auto-Encoder                                                   & ANN       &       &       & 0.742 \\
\cite{vanzyl_remove_2024}     & Auto-Encoder                                                   & ANN       & 0.989 & 0.748 & 0.660 \\
\cite{pritchard_rfi_2024}     & \begin{tabular}[c]{@{}l@{}}Auto-Encoder\end{tabular} & ANN2SNN       & 0.609 & 0.321 & 0.408 \\
This work                     & BPTT & SNN       & 0.346& 0.604& 0.474\\ \bottomrule
\end{tabular}
\footnotetext{Abbreviations: Divisive Normalisation `DN'; Spiking Neural Network `SNN'; Artificial Neural Network `ANN'; Backpropagation Through Time `BPTT'; Area Under the Receiver Operator Curve `AUROC', and Area Under the Precision Recall Curve `AUPRC'. While performance still lags the state of the art, these results encourage further development considering the compact size of our BPTT trained SNNs.}
\end{table}

\FloatBarrier\clearpage
\section*{Discussion}
This study demonstrates that SNNs can effectively perform time-series segmentation tasks, specifically RFI detection in radio astronomy.
Reformulating RFI detection as a time-series segmentation problem, we leveraged the temporal dynamics inherent to spiking neurons to recognise complex patterns.
We explored various spike encoding methods, including latency, rate, delta-modulation, delta-exposure and step-forward encodings to convert `visibility' spectrographic data to spikes.
We additionally explored optimising network depth and hidden layer widths alongside the spiking neuron and encoding parameters.
Our experiments show that simple feed-forward SNNs of first-order leaky integrate and fire (LiF) neurons perform well on synthetic datasets, achieving near state-of-the-art results and SNNs of second-order LiF neurons perform moderately well on real-world data from the LOFAR instrument, despite increased noise and variability, outperforming equivalent ANNs and previous ANN-to-SNN based work.
To our knowledge, this work is the first to train SNNs on a real radio astronomy observation dataset successfully.
The performance discrepancy between the two neuron types is significant, with the simpler neuron performing well on the simpler dataset, and vice-versa for the real dataset.
There is a curious overlap between the more biologically accurate neuron model performing uniquely better on the real dataset, handling noise more adeptly, which the simpler model might not capture as easily.
While these results provide an initial baseline for future work to compare against, the significant gap between synthetic and real dataset performance highlights the limits of a directly supervised approach to flagging with SNNs.
Our findings suggest latency with divisive normalisation and step-forward (direct exposure) encoding methods performed particularly well.
This indicates that SNNs may tolerate several different spike encoding schemes.

A key contribution of this work is introducing a divisive normalisation-inspired pre-processing step.
Our results indicate that this pre-processing can significantly improve detection performance, particularly on the synthetic dataset.
However, expanding the normalisation kernel size or incorporating information from additional prior time steps may further improve performance on realistic datasets, where background information is more dynamic.
With additional investigation into the effects of this pre-processing method, we uncover that this divisive normalisation step has a different impact on the simulated data to the real dataset, where supervision masks are less closely aligned to the underlying RFI.
This is turn suggests adapting the lessons learned in training SNNs on realistic radio astronomy data into other formulations of the RFI detection task, such as an anomaly detection task.

Despite promising initial results, our work has limitations.
The training procedure on the LOFAR dataset was equivalent to the synthetic HERA dataset, whereas training with different schedules, for extended time, with larger or more complex networks, or even a simulation-to-real approach, may yield additional performance.
Directly scaling this work's methods is yet to be exhausted. Still, the combination of increased computational requirements in simulating large SNNs and the wider configurability of spiking neurons makes exploring this space particularly challenging.
Further developments on both the data-preparation and architectural fronts may provide additional performance. For example, capturing the polarity information from the original observational data, developing more sophisticated architectures that consume entire spectrograms in a multi-headed approach, incorporating built-in de-noising capabilities, convolutional feature extraction, spiking LSTM networks and exploring training encodings are all of future interest.
For example, SNNs have shown promise in audio classification tasks for instance \cite{bos_sub-mw_2022} where the delays between neurons play a significant role in incorporating long-range temporal information.
Moreover, to address the difficulty in finding appropriate supervised training data in radio astronomy, finding self-supervised methods \cite{mesarcik_learning_2022} and adapting those for direct SNN training would potentially be very powerful.
Future work could also focus on integrating more sophisticated neuron models, such as sigma-delta or second-order LiF neurons, decoding the generally noisy output spike-trains in a trainable fashion, or casting other spectrographic segmentation tasks as time-series segmentation, such as oceanography or seismic inversion.

Modern radio astronomy is as much a science based on building precision instruments as high-performance computing.
Over multi-decade project horizons, integrating cutting-edge computing methods to find performance and efficiency benefits directly translates into greater scientific impact over the life of a telescope.
Despite this work's promising first steps, significant challenges remain regarding the operational integration of neuromorphic computing into radio astronomy processing.
While access to commercial nascent neuromorphic hardware has never been easier, most platforms focus primarily on image-oriented \cite{davies_loihi_2018, modha_neural_2023} or narrow-width time-series inputs (relative to the needs of radio astronomy) \cite{synsense_ag_xylo_2022}.
Platforms like Intel's Loihi \cite{davies_loihi_2018, noauthor_intel_nodate} provide greater flexibility, but the practical performance and efficiency of such systems on RFI detection is yet to be known.
Moreover, many focus on edge-computing use cases, where small size, weight, and power consumption dominate design considerations.
Nevertheless, we discuss preliminary feasibility and efficiency estimates in the methods section, finding RFI detection as time-series segmentation feasible on the order of \textless50 mW per baseline with contemporary hardware.
In an observatory environment, the opportunity exists to target physically larger, more powerful neuromorphic systems that still provide energy efficiencies relative to contemporary high-performance computing systems.
Moreover, a practical claim of RFI detection techniques suitable for real-time processing demands rigorous performance analysis that may impose strict requirements on exposure times, for example, when considering the time-series segmentation method this article presents.
Integrating neuromorphic computing into radio astronomy or other data-intensive sciences could be a prime candidate for hardware-software co-design.
Finally, there exists the challenge, not unique to potential SNN-based methods, of integrating machine-learned or automated methods, which can be brittle and difficult to adapt quickly, into operational observatories generally.

In conclusion, our investigation continues to pave the way for applying SNNs and neuromorphic computing in radio astronomy, potentially offering unique advantages over contemporary methods. We contribute to advancing efficient, adaptive data processing techniques essential to the future of radio astronomy while simultaneously finding a new domain to explore the capabilities of neuromorphic computing and SNNs.
\section*{Methods}
\subsection*{HERA Dataset}
The synthetic dataset Mesarcik et al. \cite{mesarcik_learning_2022} describe fully comes from the Hydrogen Epoch of Reionisation Array (HERA) simulator \cite{deboer_hydrogen_2017}.
This dataset contains 420 training spectrograms and 140 test spectrograms.
Each spectrogram covers a 30-minute simulation integrating every 3.52 seconds (producing 512 time steps) and 512 frequency channels from 105MHz to 195MHz.
The RFI is synthetic but modelled from satellite communications (localised in frequency, covering all time), lightning (covering all frequencies, localised in time), ground communication (localised in frequency, partially localised in time), and impulse blips (localised in frequency and time).
Overall, per-pixel RFI contamination sits at 2.76\%.
\subsection*{LOFAR Dataset}
The LOFAR dataset, as fully described by Mesracik et al. \cite{mesarcik_learning_2022}, consists of 7,500 training spectrograms with corresponding masks produced by AOFlagger \cite{offringa_aoflagger_2010}, along with 109 testing spectrograms that have been expertly hand-labelled.
This dataset originates from five observations, totalling 1.7 TB in size. To reduce this volume, the dataset was downsampled by selecting only the first Stokes parameter and randomly sampling 1,500 baselines, resulting in a more manageable size of approximately 10 GB.
Combining machine-generated labels and hand-labelled testing examples adds to the complexity and challenge of the dataset in addition to a significantly more difficult noise environment.
We trained on 15\% of the training samples for our experiments, 1125 spectrograms.
The motivation for selecting both this dataset and the HERA simulation dataset is that other methods have been benchmarked on these datasets, which provides a convenient way to compare the performance of our methods against traditional algorithmic approaches and other machine learning based RFI detection schemes.
\subsection*{SNN Neuron Models}
This work utilises two neuron types, a first-order leaky integrate and fire neuron model reset with subtraction, and a second-order leaky integrate and fire model, both found in snnTorch \cite{eshraghian_training_2023}.
For the first-order model
The linear differential equation modelling the membrane potential $U(t)$ dynamics of a single neuron subject to an input current $I_{in}$ is given as:
\begin{equation}
        \tau\frac{dU(t)}{dt} = -U(t) + I_{in}(t)R,
\end{equation}
where $\tau = RC$ is a fixed time constant.
In practice, the recurrent behaviours of the neuron are given as follows:
\begin{equation}
    U[t+1] = \beta U[t] + I_{in}[t+1] - RU_{thr},
\end{equation}
where $\beta$ is a tunable decay-rate parameter, $U_{thr}$ is the neuron's membrane threshold, and $R$ is a gating parameter subtracting the threshold potential if a spike is emitted in the previous time step.

A second-order LiF model adds an additional decay term, modelling both a membrane and synaptic state, replacing $I_{in}$ with $I_{syn}$ modelled as:
\begin{equation}
    I_{syn}[t+1] = \alpha I_{syn}[t] + I_{in}[t+1]
\end{equation}
which adds an additional parameter $\alpha$ which we initialize to the same values as $\beta$, since both are adjusted during training.
The second stateful element allows in principle for learning more nuanced temporal relationships, but is also more difficult to train.
We provide results for both datasets using both neuron models, reporting on the best results in the main text and adding the remaining results in Supplementary Table 1 for the HERA dataset and Supplementary Table 2 for the LOFAR dataset.
\subsection*{Simulation environment}
All simulations were conducted on the Setonix supercomputer, an HPE Cray EX system located at the Pawsey Supercomputing Centre in Western Australia. Specifically, each trial utilised a single node with a single AMD EPYC 7A53 `Trento' 64-core CPU, eight AMD Instinct MI250X GPUs, and 256 GB of RAM.
For SNN network training, we employed the snnTorch framework \cite{eshraghian_training_2023} (v0.9.0), with distributed training managed by the Lightning library \cite{falcon_pytorch_2019} (v2.2.0) to efficiently leverage the available GPU resources.
\subsection*{Hyper-Parameter tuning details}
We use the tree-structured Parzan estimation algorithm Optuna \cite{akiba_optuna_2019} exposes for multi-variate optimisation.
We simultaneously optimised for per-pixel accuracy, Area Under the Receiver Operating Characteristic curve (AUROC), Area Under the Precision-Recall Curve (AUPRC), Accuracy, and F1-Score.
AUROC evaluates the ratio of True Positive Rate (TPR) and False Positive Rate (FPR) across all possible classification thresholds.
AUPRC gives the ratio of precision and recall across several thresholds, in this case referring to the proportion of correctly classified RFI over all RFI predictions. Recall, in this case, is the TPR.
Accuracy is the per-pixel output accuracy. We expect accuracy to be very high for both the synthetic and real radio astronomy datasets as true RFI flags are very sparse; a completely silent network would still be more than 90\% accurate.
F1-Score is the harmonic mean of precision and recall at a given threshold.
The sparsity of RFI in this dataset means we generally expect high AUROC and high accuracy values with lower AUPRC and F1 scores. A good-performing model must be silent most of the time while providing confident RFI detections, hence our inclusion of both raw accuracy and F1-scores.
Results for each trial were uploaded to a persistent database from which Optuna read, allowing each trial to execute as an individual job.
Each hyper-parameter trial ran for 50 epochs with an initial learning rate of $1e-3$. However, utilising Lightning's plateau scheduler reduced the learning rate by half for every ten epochs with no improvement in validation loss.
Batch size for all trials was fixed at $36$.
Supplementary Table 3 contains the ranges and descriptions of each hyper-parameter tuned for each encoding.
For all encoding methods, we took the trial with the highest-scoring results in the most categories into repeated trials, tie-breaking by F1-Score performance.
\subsection*{Repeat trial simulation details}
Repeat trial simulations were run for 100 epochs, each with the same learning rate scheduling and batch size as the hyper-parameter trials.
Since the HERA and LOFAR datasets are of different sizes, however, despite the same number of epochs, the LOFAR trials are subject to more gradient updates per epoch, and due to the longer exposure times, the SNNs are evolved for a longer time proportional to this exposure parameter. 
\subsection*{Spike Encoding Methods}
Each encoding method involves a spike-encoding scheme, decoding scheme and loss function.
Figure \ref{fig:encodings} contains example plots for all encodings with a sample spectrogram exposed for four time steps where applicable.
As mentioned previously, these encoding methods take a set of `visibilities' $V(\upsilon, T, b)$ and output a spike-train with dimensions $V(\upsilon, T \cdot E, b)$ where $E$ is an integer exposure time.
\subsubsection*{Latency}
Our latency encoding scheme is relatively straightforward. For a given exposure $E$, input pixel intensities are mapped inversely linearly from 0 to $E$, meaning high-intensity pixels spike almost immediately and low-intensity pixels progressively later.

For output decoding, we interpret any spikes before the last exposure step as an RFI mask and no spikes or a spike in the final exposure as background. The following equation maps the spike times in the supervised masks, that is, the $t$ value in $F(\upsilon, t, b)$:
\begin{equation}\label{eq:latency:mask}
    t = \begin{cases}
        0 & G(\upsilon, t, b) = 1 \\
        E & otherwise
    \end{cases}
\end{equation}
The comparison function, $\mathcal{H}$, is the mean square spike time governed by the following function:
\begin{equation}\label{eq:latency:comparison}
    \mathcal{H}_{latency} = \Sigma_e^E\Sigma_\upsilon^\Upsilon(y_{\upsilon,e} - f_{\upsilon, e})^2.
\end{equation}
Figure \ref{fig:encodings}.b contains an example input raster plot.
\subsubsection*{Rate}
Rate encoding is well-understood and widely used when processing image data in SNNs. For a given exposure $E$, we interpret pixel values as firing probabilities. We map supervised masks to firing rates of 0.8 for pixels labelled as containing RFI and 0.2 for pixels that do not. For output decoding, we interpret outputs with a firing rate greater than 0.75 as an RFI flag.

To map rate-encoding into a time-series segmentation problem, we treat each time step in the original signal as a classification problem. The comparison function, $\mathcal{H}$, is an adjusted mean square error spike count loss, adapted from the implementation in snnTorch \cite{eshraghian_training_2023} to handle time segments with no positive labels present, and is governed by the following equation:
\begin{equation}
    \mathcal{H}_{rate} = \Sigma_e^E(y_{\upsilon,e} - f_{\upsilon, e})^2.
\end{equation}
Figure \ref{fig:encodings}.c contains an example input raster plot.
\subsubsection*{Delta-Modulation}
Delta-modulation encoding encodes inputs only on relative change. Our method utilises the delta-modulation encoding method in snnTorch \cite{eshraghian_training_2023}, using both positive and negative polarity spikes. Output decoding is a more complex procedure.
Encoding regions of an RFI mask that can span as little as a single pixel requires spike polarity for `on' and `off' spikes. However, in our networks, neurons only emit positive spikes. Therefore, we double the frequency width of the output layer and move all negative polarity spikes to the lower region as positive spikes. 

The comparison function, $\mathcal{H}$, is the Huber loss function, which combines L1 and MSE loss functions, computing the squared MSE term if the error falls below a threshold and L1 loss otherwise. We employ this loss function to handle the sparsity of the output spike trains. The Huber loss as acting on spike-trains is:
\begin{equation}
    \mathcal{H}_{delta} = \begin{cases}
        \frac{1}{2}(y - f)^2 & |y - f| \leq \delta \\
        \delta (|y - f| - \frac{1}{2}\delta) & otherwise
    \end{cases}
\end{equation}
Figure \ref{fig:encodings}.d contains an example input raster plot.
\subsubsection*{Delta-Exposure}
Delta exposure is a hybrid of the latency and delta-modulation methods.
Input encoding is similar to delta modulation.
However, we include an exposure time to give the network more time to react to the input stimulus.
Output decoding is also simplified to the scheme described in equation \ref{eq:latency:comparison}.
Figure \ref{fig:encodings}.e contains an example raster for this scheme.

\subsubsection*{Step-Forward}
\begin{figure*}[!htbp]
    \centering
        \begin{subfigure}{0.45\columnwidth}
        \centering
        \includegraphics[width=\textwidth,  keepaspectratio]{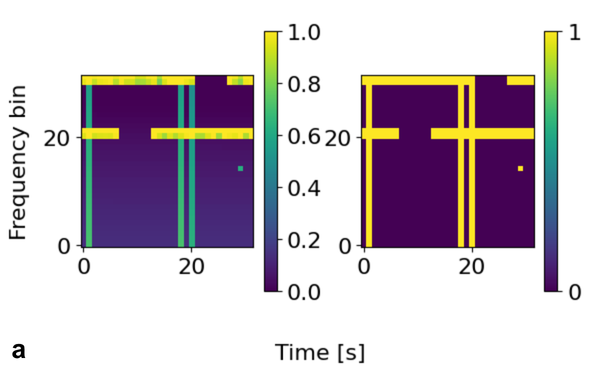}
        \label{fig:enc:original}
    \end{subfigure}
    \begin{subfigure}{0.45\columnwidth}
        \centering
        \includegraphics[width=\textwidth,  keepaspectratio]{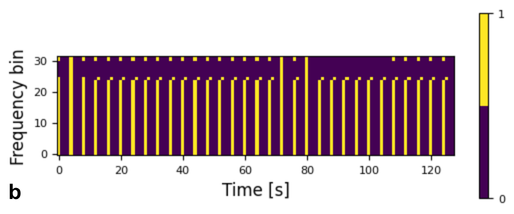}
        \label{fig:enc:latency}
    \end{subfigure}
    \begin{subfigure}{0.45\columnwidth}
        \centering
        \includegraphics[width=\textwidth,  keepaspectratio]{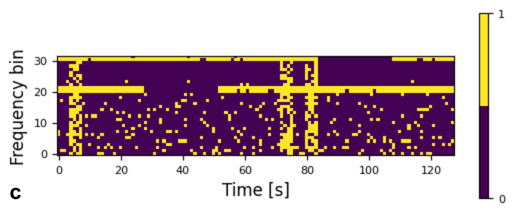}
        \label{fig:enc:rate}
    \end{subfigure}
    \begin{subfigure}{0.45\columnwidth}
        \centering
        \includegraphics[width=0.6\textwidth,keepaspectratio]{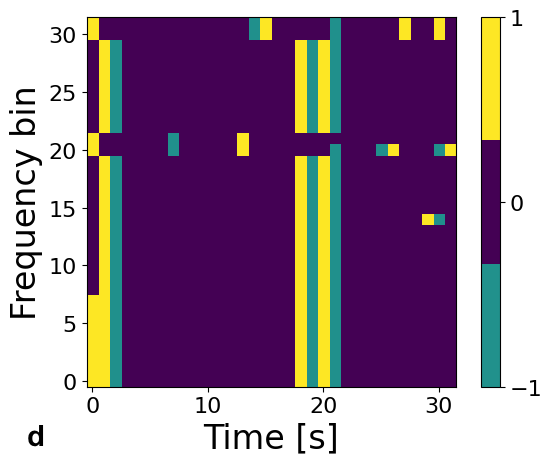}
        \label{fig:enc:delta}
    \end{subfigure}
        \begin{subfigure}{0.45\columnwidth}
        \centering
        \includegraphics[width=\textwidth,  keepaspectratio]{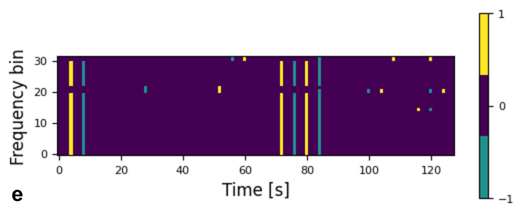}
        \label{fig:enc:delta_exposure}
    \end{subfigure}
    \begin{subfigure}{0.45\columnwidth}
        \centering
        \includegraphics[width=0.6\textwidth,  keepaspectratio]{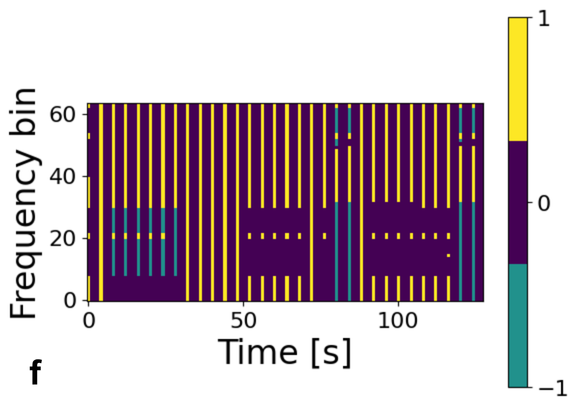}
        \label{fig:enc:forwardstep:first}
    \end{subfigure}
    \begin{subfigure}{0.45\columnwidth}
        \centering
        \includegraphics[width=0.6\textwidth,  keepaspectratio]{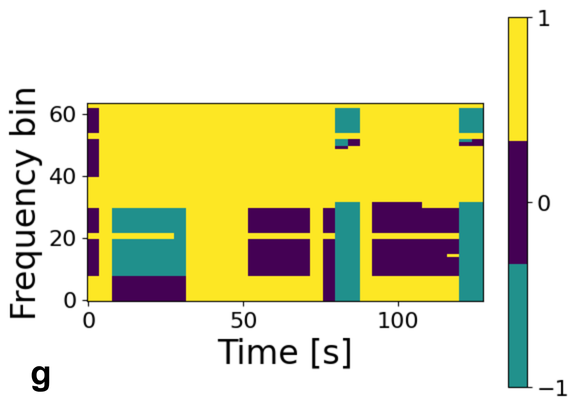}
        \label{fig:enc:forwardstep:direct}
    \end{subfigure}
    \begin{subfigure}{0.45\columnwidth}
        \centering
        \includegraphics[width=0.6\textwidth,  keepaspectratio]{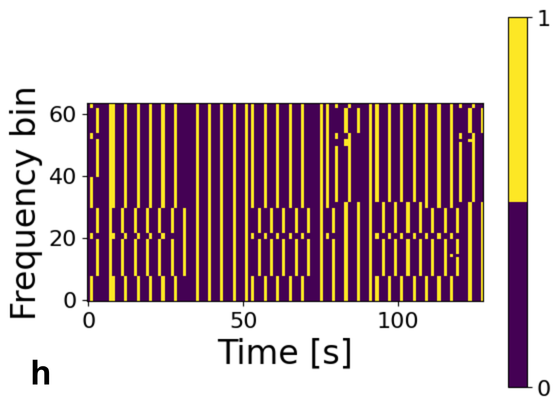}
        \label{fig:enc:forwardstep:latency}
    \end{subfigure}
    \caption{Spike Raster Plots for Various Encodings Applied to Synthetic HERA Data. Encodings are exposed for four time steps where applicable, showcasing how different methods capture distinct Radio Frequency Interference (RFI) features. Sub-panels depict: (a) original input spectrogram and associated RFI mask; (b) latency encoding reveals clear banding and blip features; (c) rate encoding reveals most features clearly but struggles with isolated blips; (d) delta encoding condenses frequency-isolated RFI into a few spikes, making thresholding during encoding critical; (e) delta exposure encoding improves detail but still struggles with frequency-isolated RFI; (f) step-forward encoding with `first' exposure mode clearly shows banding but is hard to interpret; (g) step-forward encoding with `direct' exposure is active but shows the difficulty in identifying bands, and (h) step-forward encoding example with `latency' exposure effectively identifies bands and blips.}
\label{fig:encodings}
\end{figure*}

Visibility data is spatio-temporal, so we tested encoding methods based on encoding audio data, the Step-Forward algorithm \cite{stewart_speech2spikes_2023}. This approach takes the step-wise difference between the signal at each time step and a running cumulative sum, emitting a positive or negative spike when the difference crosses a threshold (in our case, $0.1$) upwards or downwards. Additionally, a cumulative sum of these differences is appended to the end of the frequency channels, doubling the width of the input. Output decoding and supervised mask encoding, covered in Equation \ref{eq:latency:mask}, are handled identically to latency encoding. The comparison function is also identical to latency encoding, covered in Equation \ref{eq:latency:comparison}. We provide three methods to adapt the original step-forward algorithm to the time-series segmentation problem outlined below; each method handles the required exposure parameter.
\paragraph{First}
This exposure mode presents spikes to the network only on the first exposure step for each time step, with silence following.
Figure \ref{fig:encodings}.f contains an example raster plot for this exposure method.
\paragraph{Direct}
This exposure mode presents spikes to the network at every exposure step for each time step.
Figure \ref{fig:encodings}.g contains an example raster plot for this exposure method.
\paragraph{Latency}
This exposure mode presents spikes to the network latency encoded at exposure time $0$ where a spike is present and exposure time $E$ otherwise.
Figure \ref{fig:encodings}.h contains an example raster plot for this exposure method.
\subsection*{Dataset Preprocessing}
Here we discuss our preprocessing methods in more detail and highlight the differences between the simulated HERA and observed LOFAR datasets.
Specifically, by highlighting these differences, we aim to at least partially explain the discrepancy in task performance of SNNs trained to perform RFI detection in a supervised manner.
Aside from divisive normalisation, we preprocess the HERA dataset by performing min-max scaling between values one standard deviation below the mean and four standard deviations above the mean.
We preprocess the LOFAR dataset by performing min-max scaling between values three standard deviations below the mean and 95 standard deviations above the mean.
Both datasets are scaled in this manner before applying divisive normalisation.
We chose these methods and values based on prior literature \cite{mesarcik_learning_2022}.

Table \ref{tab:preprocess} contains the maximum, mean, and median values of the pixels classified as noisy and noiseless in each dataset, both with and without divisive normalisation and under an older and revised pre-processing pipeline for the LOFAR dataset.
\begin{sidewaystable}[!htbp]
\caption{Summary statistics between noisy and noiseless pixels in the HERA and LOFAR datasets with and without Divisive Normalisation (DN).}
\label{tab:preprocess}
\begin{tabular}{@{}llllllllll@{}}
\toprule
Dataset            & \multicolumn{1}{c}{\begin{tabular}[c]{@{}c@{}}Max\\ (noisy)\end{tabular}} & \multicolumn{1}{c}{\begin{tabular}[c]{@{}c@{}}Max\\ (noiseless)\end{tabular}} & \multicolumn{1}{c}{Difference} & \multicolumn{1}{c}{\begin{tabular}[c]{@{}c@{}}Mean\\ (noisy)\end{tabular}} & \multicolumn{1}{c}{\begin{tabular}[c]{@{}c@{}}Mean\\ (noiseless)\end{tabular}} & \multicolumn{1}{c}{Difference} & \multicolumn{1}{c}{\begin{tabular}[c]{@{}c@{}}Median\\ (noisy)\end{tabular}} & \multicolumn{1}{c}{\begin{tabular}[c]{@{}c@{}}Median\\ (noiseless)\end{tabular}} & \multicolumn{1}{c}{Difference} \\ \midrule
HERA - Train       & 1.00                                                                      & 0.8955                                                                        & 0.1045                         & 0.5423                                                                     & 1.00                                                                           & 0.4577                         & 0.5904                                                                       & 0.5904                                                                           & 0.00                           \\
HERA - Test        & 1.00                                                                      & 0.8742                                                                        & 0.1258                         & 0.6616                                                                     & 0.5349                                                                         & 0.1267                         & 0.7221                                                                       & 0.5963                                                                           & 0.1258                         \\
HERA - DN - Train  & 1.00                                                                      & 0.1179                                                                        & 0.8821                         & 0.2670                                                                     & 0.00                                                                           & 0.2670                         & 0.2370                                                                       & 0.00                                                                             & 0.2370                         \\
HERA - DN - Test   & 1.00                                                                      & 0.0996                                                                        & 0.9004                         & 0.2579                                                                     & 0.00                                                                           & 0.2579                         & 0.2207                                                                       & 0.00                                                                             & 0.2207                         \\
LOFAR - Train      & 1.00                                                                      & 0.7612                                                                        & 0.2388                         & 0.2226                                                                     & 0.0434                                                                         & 0.1792                         & 0.1424                                                                       & 0.00                                                                             & 0.1424                         \\
LOFAR - Test       & 1.00                                                                      & 1.00                                                                          & 0.00                           & 0.3926                                                                     & 0.0087                                                                         & 0.3839                         & 0.3039                                                                       & 0.00                                                                             & 0.3039                         \\
LOFAR - DN - Train & 0.9865                                                                    & 0.9865                                                                        & 0.00                           & 0.0623                                                                     & 0.0105                                                                         & 0.0518                         & 0.00                                                                         & 0.00                                                                             & 0.00                           \\
LOFAR - DN - Test  & 0.9921                                                                    & 0.9921                                                                        & 0.00                           & 0.1187                                                                     & 0.0068                                                                         & 0.1119                         & 0.00                                                                         & 0.00                                                                             & 0.00                           \\ \bottomrule
\end{tabular}
\end{sidewaystable}
We see that in the HERA dataset there are significant differences in the maximum values in the noiseless and noisy supervision masks which are extended by the divisive normalisation technique.
Whereas in the LOFAR dataset, this is not the case, in fact, there is no difference at all.
In the LOFAR dataset, supervision masks are provided as bounding areas which are not as tightly bound to the RFI as in the simulated, and therefore perfect, supervision masks.
Moreover, the differences between the train and test sets for the HERA dataset are generally closer than those found in the LOFAR dataset which suggests that the behaviours learned in the training set may apply to the test set more readily than those discovered on the LOFAR training set.

\subsection*{Patching and Banding}
Patching, where we split each spectrogram into $32 \times 32$ sized sections for individual inference is necessary to make training practical under backpropagation through time.
This approach is a well known approach to dealing with radio astronomy spectrographic data \cite{mesarcik_learning_2022} and while convenient, this does mean the same SNN model is exposed to all frequency channels during training.
In some trials the model learns a bias to flag particular (relative) frequency inputs or time-steps consistently, leading to visible banding in the output.
Once reconstructed, this appears as regular banding across an entire spectrogram.
Another cause of a similar error is a quirk of the latency encoding and decoding method.
When decoding specifically, any output (for a single spectrogram time-step) over an exposure time occurring before the final time-step is considered RFI, and the model is trained to output spikes at the last time-step to indicate no RFI (having spikes on all outputs is necessary to allow backpropagation to be effective, otherwise the neurons `die’ and cease responding to any inputs).
Where the network lacks confidence, it can output spikes too early which contributes to the consistent lines across patch boundaries.
Supplementary Figure 1 contains example inference of this phenomena.
To encourage the SNNs to avoid learning aggressive biases, hyper-parameters such as neuron thresholds and decay rates, in addition to learning rates helps ameliorate this issue.
Ideally, a separate model would be trained on a consistent section of the frequency spectrum. 
\subsection*{Computational and Energy Efficiency}
While not the main focus of this work, we remark briefly on the potential computational and energy efficiency of our proposed method to provide intuitive operational motivation behind exploring SNNs for RFI detection.
Specifically, based on available spec-sheet data, we provide a lower and upper estimate on power consumption per baseline for SynSense Xylo hardware \cite{synsense_ag_xylo_2022}.
We additionally reason about the clock speeds required for real-time processing of RFI under our time-series segmentation approach, which has very different ergonomics to image-based processing solutions.
Finally, we compare the number of parameters in our models to those of other state-of-the-art solutions, to provide intuition about what kind of practical difference a move to neuromorphic computing could make.

SynSense Xylo 2 is a commercially available neuromorphic chipset designed mainly for audio-signal processing with SNNs.
Many other neuromorphic computing chipsets exist, however, we choose to focus on SynSense owing to readily available specification data. 
The SynSense Xylo 2 chipset has eight input channels, a maximum of 1024 hidden neurons, a minimum power draw of $216\mu$W and a maximum power draw of $550\mu$W \cite{noauthor_xylo-audio_2022}.
Processing a single baseline would require 64 Xylo chipsets and consume between 13.8 and 35.2 mW of power.
The limited bandwidth and hidden neuron count of this hardware preclude an image-based approach to RFI detection, and it is for this reason that a time-series segmentation approach is explicitly required for RFI detection in available neuromorphic hardware.
A more thorough follow-up study into the practical deployment of neuromorphic computing in radio astronomy observatories is critical; this initial effort to validate the overall approach is a necessary first step.

Regarding real-time suitability, RFI detection would need to occur within the integration time of the instrument itself; processing a spectrographic time-step in $\approx$ 3.52 seconds (in the case of the HERA simulation) \cite{mesarcik_learning_2022-1}.
Even with 64 exposure time-steps, the maximum we consider in this work, and a very conservative assumption of one million operations per exposure, this places a minimum clock speed of 32MHz, which is significantly lower than the listed clock rate of 50MHz \cite{noauthor_xylo-audio_2022}.
Again, the point here is not to make a concrete claim around actual real-time performance, but to indicate that a follow-up study with real neuromorphic hardware is deserved and likely to yield positive results.

Finally, to remark on model size, our HERA models contain 73.7k and 203k parameters for the original and normalised datasets, respectively, while our LOFAR models contain 32.8k and 16.4k parameters, respectively.
The counter-intuitive observation that the models for the more complex dataset are smaller highlights another challenge in utilising SNNs; supervised training requires error to propagate through time.
Where the supervising signal is sparse, as in the LOFAR dataset, it makes learning difficult; however, each parameter in a smaller network will receive relatively more supervision signal in each pass.
Even our largest models, compared to recent studies into the computational requirements of deep learning methods in RFI detection \cite{dutoit_comparison_2024}, where UNet-style models use up to 800k parameters, are significantly smaller. 
This difference highlights the key driving benefit of moving to a time-series approach to RFI detection with SNNs; the input data rate is constrained to the number of instrument channels, and the time axis is pushed into the network itself.

While further work is required to properly characterise the computational and energy efficiency of a neuromorphic approach to RFI detection, this work provides a precedent for further investigation.
It suggests that such a method could provide significant operational benefits.
\section*{Data availability}
This study's original datasets are openly available online \cite{mesarcik_learning_2022-1} under a CC BY 4.0 license.
\section*{Code availability}
Source code is openly available \cite{pritchard_pritchardnsnn-rfi-super_2024} under an MIT license.
\bibliography{NatComms2024}
\section*{Acknowledgments}
This work was supported by a Westpac Future Leaders Scholarship, an Australian Government Research Training Program Fees Offset and Stipend, and resources provided by the Pawsey Supercomputing Research Centre with funding from the Australian Government and the Government of Western Australia.
\section*{Author contributions}
N.J.P., A.W., M.B., and R.D. all contributed conceptually.
N.J.P. and A.W. conceived the idea.
N.J.P. designed and performed the simulations.
N.J.P., A.W., M.B., and R.D. wrote the manuscript.
\section*{Competing interests}
The authors declare no competing interests.
\section*{Additional Information}
\paragraph{Correspondence} and requests for materials should be addressed to Nicholas J. Pritchard.
\newpage
\renewcommand{\tablename}{Supplementary Table}
\renewcommand{\figurename}{Supplementary Figure}
\setcounter{figure}{0}
\setcounter{table}{0}
\section*{Supplementary Note 1: HERA Results with Second-Order LiF Neurons}
\begin{table}[!htbp]
\centering
\caption{These results show the performance of each encoding method using the final hyper-parameters on the HERA dataset with second-order LiF neurons.}
\label{tab:hera-trials}
\begin{tabularx}{\linewidth}{@{}ccccccccc@{}}
\toprule
Encoding Method               & \multicolumn{2}{c}{Accuracy} & \multicolumn{2}{c}{AUROC} & \multicolumn{2}{c}{AUPRC} & \multicolumn{2}{c}{F1} \\ \midrule
Delta Exposure                & 0.958             & 0.017    & 0.645            & 0.147  & 0.500            & 0.150  & 0.289          & 0.216          \\
Delta Exposure + DN      & 0.962             & 0.010    & 0.628            & 0.159  & 0.561            & 0.090  & 0.270          & 0.254          \\
Latency                       & 0.966             & 0.008    & 0.840            & 0.176  & 0.688            & 0.112  & 0.536          & 0.243          \\
Latency + DN    & 0.986    & 0.008    & 0.898            & 0.046  & 0.795            & 0.092  & 0.784 & 0.097          \\
Step-Forward-Direct           & 0.970             & 0.005    & 0.704   & 0.087  & 0.493            & 0.071  & 0.420          & 0.136          \\
{\ul{Step-Forward-Direct}} + DN & \textbf{0.988}             & 0.004    & \textbf{0.943}            & 0.019  & \textbf{0.821}            & 0.057  & \textbf{0.789}          & 0.078          \\
\bottomrule
\end{tabularx}
\footnotetext{Each metric is listed as mean and standard deviation. Ten trials were completed for each encoding method. Divisive Normalisation abbreviated to `DN'. The best scores are bolded. Results lag those using simpler first-order LiF neurons, pointing to a fundamental difference in task complexity.}
\end{table}
Table \ref{tab:hera-trials} presents results for the HERA dataset using second-order LiF neurons.
These results lag in performance compared to the simpler first-order LiF neuron results presented in the main-text, however, divisive normalisation is more impactful.
However, some notable trends are worth mentioning.
Here, step-forward (direct) encoding is clearly the best performing encoding and divisive normalisation increases accuracy by 0.014, AUROC by 0.093, AUPRC by 0.165 and F1 by 0.199 on average across all encodings.
\newpage
\section*{Supplementary Note 2: LOFAR Results with First-Order LiF Neurons}
\begin{table}[!htbp]
\centering
\caption{The final results show the performance of each encoding method using the final hyper-parameters on the LOFAR dataset with first-order LiF neurons.}
\label{tab:lofar-trials}
\begin{tabularx}{\linewidth}{@{}cllllllll@{}}
\toprule
Encoding Method               & \multicolumn{2}{c}{Accuracy} & \multicolumn{2}{c}{AUROC} & \multicolumn{2}{c}{AUPRC} & \multicolumn{2}{c}{F1} \\ \midrule
Delta Exposure                & 0.973             & 0.007    & 0.149            & 0.009  & 0.556            & 0.002  & 0.120          & 0.017          \\
Delta Exposure + DN      & 0.961             & 0.005    & 0.143            & 0.004  & 0.555            & 0.003  & 0.097          & 0.006          \\
Latency                       & 0.972             & 0.017    & 0.190            & 0.037  & 0.566            & 0.004  & 0.149          & 0.042          \\
Latency + DN    & 0.978             & 0.003    & \textbf{0.630}   & 0.003  & 0.254            & 0.011  & 0.198          & 0.016          \\
{\ul{Step-Forward-Direct}}           & \textbf{0.983}    & 0.002    & 0.262            & 0.011  & \textbf{0.628}   & 0.005  & \textbf{0.217} & 0.014          \\
Step-Forward-Direct + DN & 0.945             & 0.011    & 0.218            & 0.014  & 0.613            & 0.015  & 0.128          & 0.011          \\
\bottomrule
\end{tabularx}
\footnotetext{Each metric is listed as mean and standard deviation. The best scores are bolded. Ten trials were completed for each encoding method. Divisive Normalisation abbreviated to `DN'.
The simpler neuron model performs significantly worse than a second-order LiF approximation for all encodings and in all metrics except accuracy.}
\end{table}
Table \ref{tab:lofar-trials} contains results on the LOFAR dataset using the simpler first-order LiF neurons.
These results lag those reported in the main-text by a significant margin, although again, step-forward (direct) encoding performs best.
\section*{Supplementary Note 3: Hyperparameter Ranges}
\begin{table}[!htbp]
\centering
\caption{Parameter ranges for attributes included in the final hyper-parameter searches.}
\label{tab:hyperparam-range}
\begin{tabular}{@{}ll@{}}
\toprule
Attribute  & Parameter Range \\ \midrule
Num Hidden & 128, 256, 512 \\
Num Layers & 2 - 6        \\
Beta       & 0.5 - 0.99      \\
Exposure   & 1 - 64          \\ \bottomrule
\end{tabular}
\footnotetext{`Num Hidden' refers to the number of neurons in each hidden layer and `Num Layers' refers to the number of hidden layers.}
\end{table}
In Supplementary Table \ref{tab:hyperparam-range}, `Num hidden' refers to the number of neurons in each hidden layer.
`Num layers' refers to the number of hidden layers in the network.
Beta defines the excitability of LiF neurons within the network, and exposure is a parameter for the encoding method used in all but the delta-modulation method. Beta and exposure hold no meaning for ANN networks and are omitted for these trials.
\section*{Supplementary Note 4: Patching and Banding Example}
\begin{figure}[!htbp]
    \centering
    \begin{subfigure}{0.45\columnwidth}
        \centering
        \includegraphics[height=1.5in, keepaspectratio]{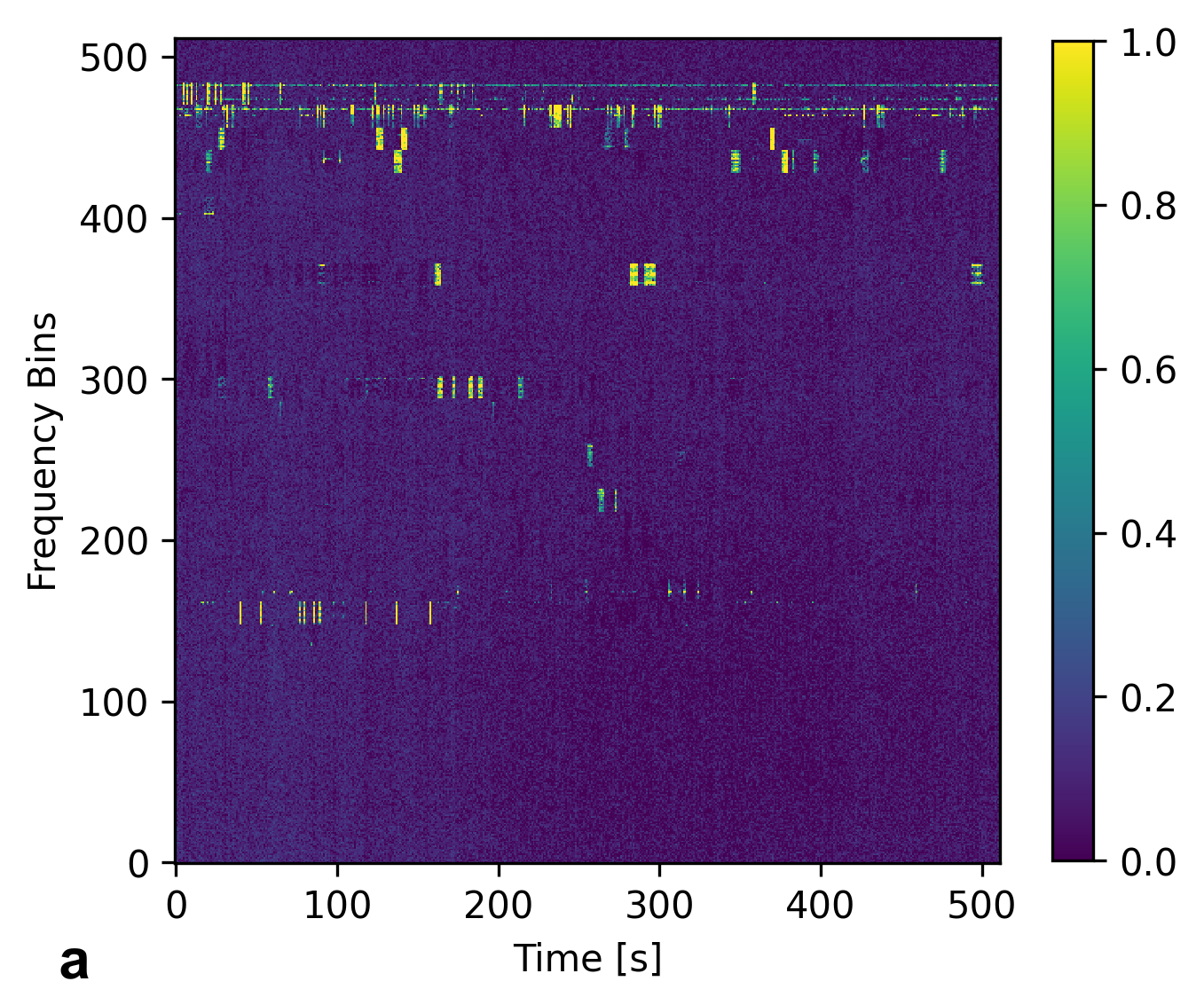}
        \label{fig:res:lofar:orig:orig}
    \end{subfigure}
    \begin{subfigure}{0.45\columnwidth}
        \centering
        \includegraphics[height=1.5in, keepaspectratio]{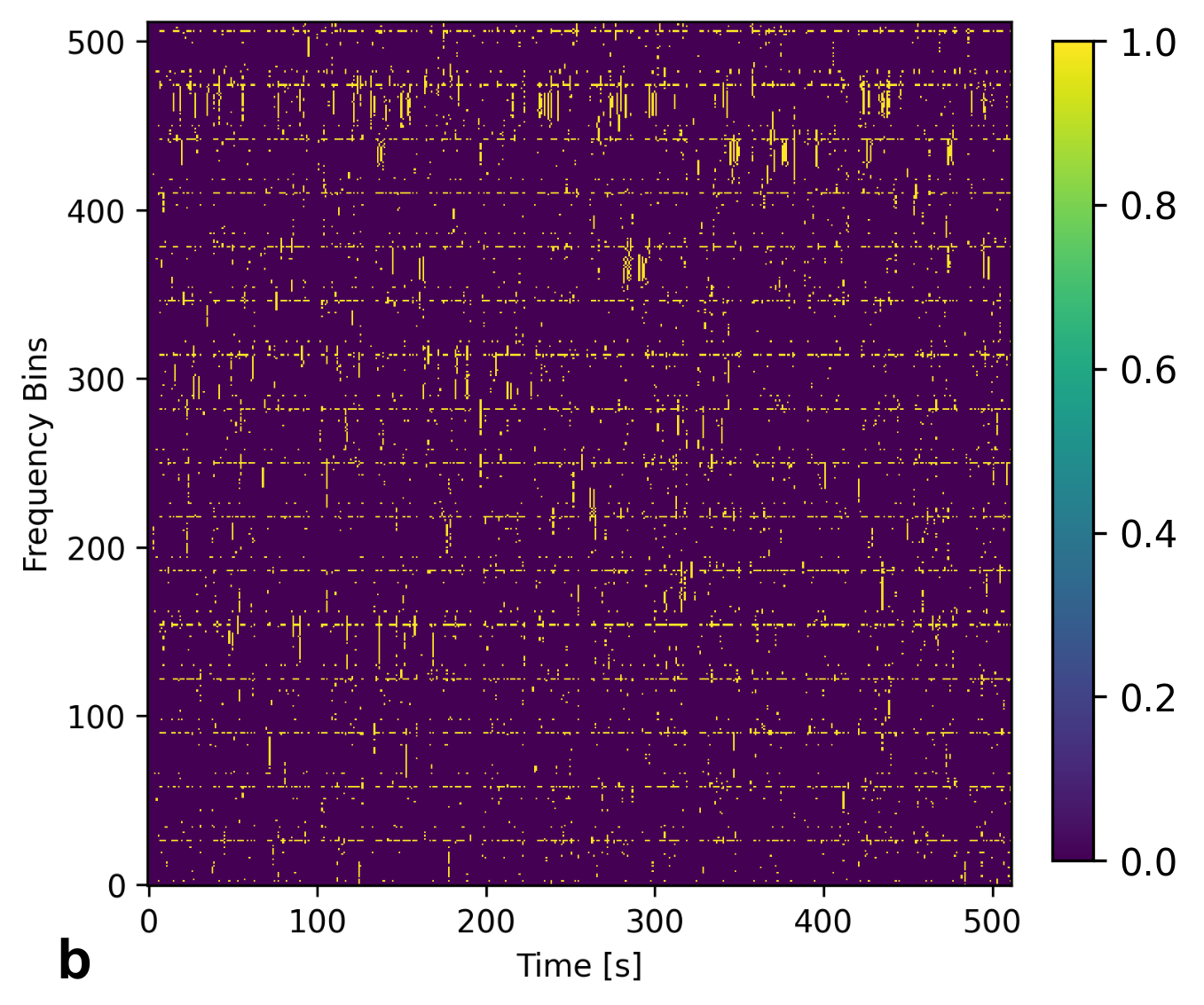}
        \label{fig:res:lofar:inf-orig:orig}
    \end{subfigure}
    \begin{subfigure}{0.45\columnwidth}
        \centering
        \includegraphics[height=1.5in, keepaspectratio]{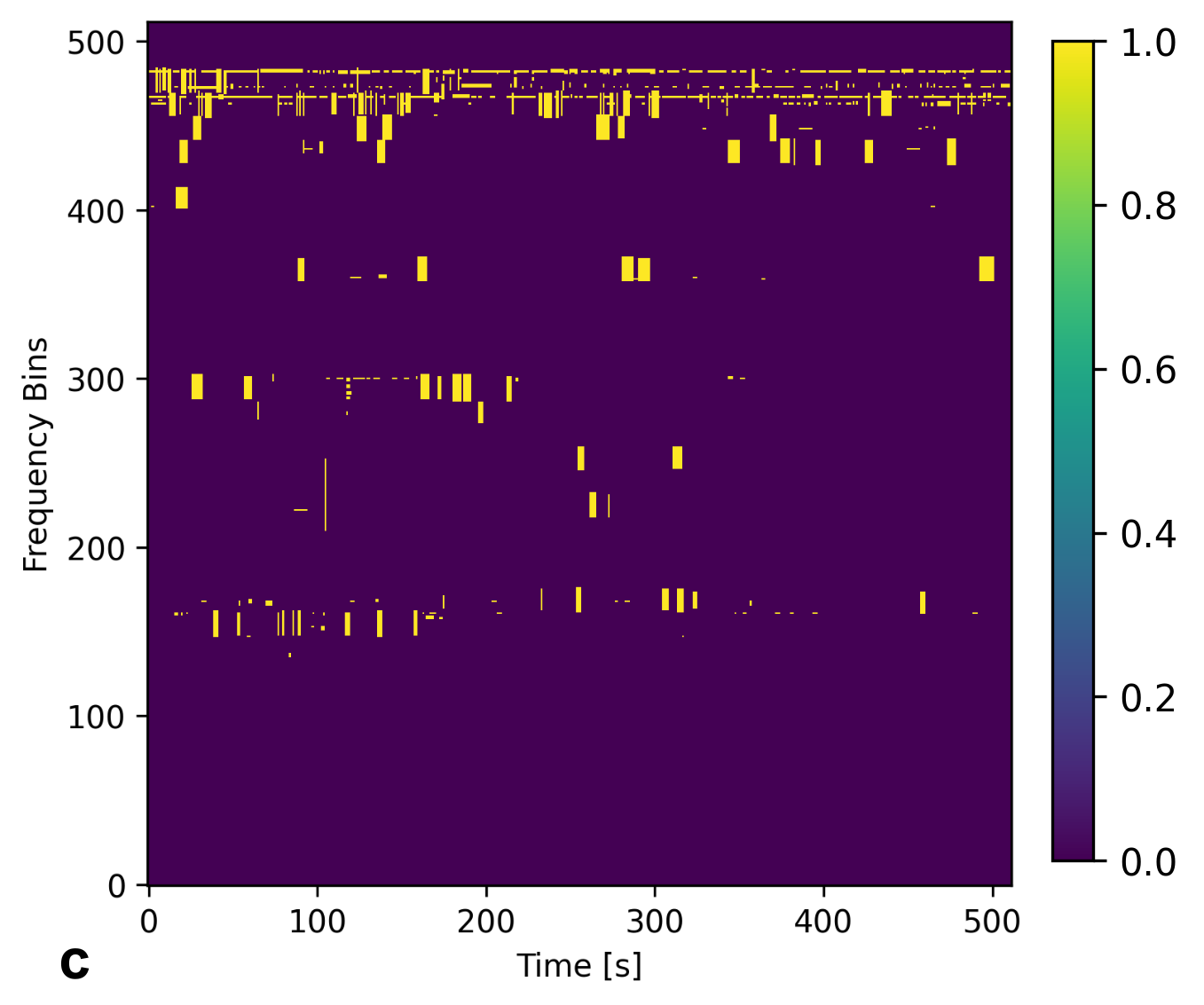}
        \label{fig:res:lofar:mask:orig}
    \end{subfigure}
    \caption{Example of banding in patched inference over an entire LOFAR spectrogram, exhibiting noticeable banding due to bias. Sub-panels depict: (a) original spectrogram; (b) latency-based inference on the original spectrogram, and (c) expert-labelled Radio Frequency Interference (RFI) mask.}
\label{fig:example:lofar-band}
\end{figure}
\end{document}